\documentclass[]{bytedance_seed}

\usepackage[toc,page,header]{appendix}

\usepackage{minitoc}

\title{\fontsize{16pt}{20pt}\selectfont CoDiQ: Test-Time Scaling for Controllable Difficult Question Generation}

\author[1,3]{Zhongyuan Peng}
\author[1,2]{Caijun Xu}
\author[1]{Changyi Xiao}
\author[1]{Shibo Hong}
\author[3]{Eli Zhang$^\dagger$}
\author[3]{Stephen Huang}
\author[1,2]{Yixin Cao$^\dagger$}

\affiliation[1]{Fudan University}
\affiliation[2]{Shanghai Innovation Institute}
\affiliation[3]{M-A-P}

\contribution[\dagger]{Corresponding authors}

\usepackage{algorithm}
\usepackage{algorithmic}
\usepackage{xcolor}
\usepackage[most]{tcolorbox}
\usepackage{colortbl}
\usepackage{adjustbox}
\usepackage{threeparttable}
\tcbuselibrary{breakable, skins}
\usepackage{longtable}
\usepackage{blindtext}
\usepackage{wrapfig}

\definecolor{color21}{RGB}{180, 230, 190}  % 柔和绿色边框
\definecolor{color22}{RGB}{240, 255, 240}  % 极淡绿色留白
\definecolor{color11}{RGB}{140, 190, 230}  % 柔和蓝色边框
\definecolor{color12}{RGB}{230, 240, 255}  % 极淡蓝色留白
\definecolor{color31}{RGB}{255, 190, 140}   % 柔和橙色边框（较明显但不刺眼）
\definecolor{color32}{RGB}{255, 225, 210}  % 稍微深一点的橙色背景（柔和但更明显）
\definecolor{color41}{RGB}{255, 245, 180}  % 更浅柔和的黄色边框
\definecolor{color42}{RGB}{255, 250, 210}  % 更浅柔和的浅黄色背景
\definecolor{color43}{RGB}{255, 255, 240}  % 极浅黄色留白，几乎白色

\definecolor{promptboxblue}{RGB}{59, 130, 246}
\definecolor{promptboxgray}{RGB}{107, 114, 128}
\definecolor{promptboxlightgray}{RGB}{243, 244, 246}
\definecolor{promptboxgreen}{RGB}{34, 197, 94}
\newtcolorbox{promptbox}[1][]{
  enhanced,
  breakable,           % 允许跨页
  break at=-20pt,      % 调整分页点
  colback=promptboxlightgray,
  colframe=promptboxblue!30,
  arc=8pt,
  boxrule=0.5pt,
  left=12pt,
  right=12pt,
  top=8pt,
  bottom=8pt,
  fonttitle=\bfseries,
  fontupper=\linespread{1.2}\selectfont,
  title=#1,
  before skip=10pt,    % 添加前间距控制
  after skip=10pt      % 添加后间距控制
}

\abstract{
Large Reasoning Models (LRMs) benefit substantially from training on challenging competition-level questions. However, existing automated question synthesis methods lack precise difficulty control, incur high computational costs, and struggle to generate competition-level questions at scale.
In this paper, we propose \textbf{CoDiQ} (\textbf{Co}ntrollable \textbf{Di}fficult \textbf{Q}uestion Generation), a novel framework enabling fine-grained difficulty control via test-time scaling while ensuring question solvability. 
Specifically, first, we identify a test-time scaling tendency (extended reasoning token budget boosts difficulty but reduces solvability) and the intrinsic properties defining the upper bound of a model’s ability to generate valid, high-difficulty questions.
Then, we develop \textbf{CoDiQ-Generator} from Qwen3-8B, which improves the upper bound of difficult question generation, making it particularly well-suited for challenging question construction.
Building on the CoDiQ framework, we build \textbf{CoDiQ-Corpus} (44K competition-grade question sequences). Human evaluations show these questions are significantly more challenging than LiveCodeBench/AIME with over 82\% solvability. Training LRMs on CoDiQ-Corpus substantially improves reasoning performance, verifying that scaling controlled-difficulty training questions enhances reasoning capabilities.
We open-source CoDiQ-Corpus, CoDiQ-Generator, and implementations to support related research.
}

\date{February 2, 2026}
\correspondence{\email{yxcao@fudan.edu.cn}, \email{zhangge.eli@bytedance.com}}
\checkdata[Project Page]{\url{https://github.com/ALEX-nlp/CoDiQ}}

\begin{document}
\maketitle

%不需要目录就注释掉 注意目录不要和第一页放在一块 要有\newpage
%\newpage
%\tableofcontents
%\newpage

\section{Introduction}
% [Paragraph 1: Background]
The rapid advancement of Large Reasoning Models (LRMs) has demonstrated remarkable capabilities in complex reasoning, with recent works achieving impressive performance on challenging benchmarks across mathematics and coding~\cite{zhao2025promptcot, manem2025sand, pei2025scalediff}.
A crucial factor driving these improvements is the availability of high-quality training and evaluation data that truly stress reasoning, yet such data are scarce. Importantly, much like scientific discovery, finding the right difficult questions can be as critical as solving them. As difficulty rises, reliable problem construction demands expert knowledge and careful validation, making purely human-driven pipelines expensive and hard to scale.

In this paper, we aim to scale the synthesis of high-difficulty questions while keeping them well-posed and solvable. 
Recent research has explored various approaches for mathematics and programming, ranging from human-in-the-loop methodologies~\cite{shah2024ai} and adversarial generation~\cite{xie2024adversarial} to iterative evolutionary strategies~\cite{ding2024unleashing, zhao2025promptcot, manem2025sand}. 

% [Paragraph 2: Challenges]
% However, pushing difficulty at scale faces three major challenges. 
% \textbf{First, Generator capacity ceiling.} A model typically struggles to generate questions that are substantially harder than what it can reliably reason about, leading to stalled progress. 
% \textbf{Second, solvability–complexity trade-off.} Forcing complexity often breaks logical consistency, producing ``fake hard'' but unsolvable or ill-defined questions. 
% \textbf{Third, difficulty definition and control.} Difficulty is not directly observable, is domain-dependent, and is difficult to standardize. Without a measurable surrogate, ``make it harder'' is neither controllable nor comparable across domains, and curriculum-style training becomes brittle.
However, pushing difficulty at scale faces three major challenges. 
First, there is a \textit{generator capacity ceiling}, where a model typically struggles to generate questions substantially harder than what it can reliably reason about, leading to stalled progress. 
Second, the \textit{solvability–complexity trade-off} implies that forcing complexity often breaks logical consistency, producing ``fake hard'' but unsolvable or ill-defined questions. 
Finally, the \textit{difficulty definition and control.} Since difficulty is neither directly observable nor standardized, ``make it harder'' becomes uncontrollable without a measurable surrogate, rendering curriculum-style training brittle.

% [Paragraph 3: Solution]
To address these challenges, we propose \textit{CoDiQ} (Controllable Difficult Question Generation), a framework that introduces test-time scaling into question generation and systematically scales difficulty through three key innovations. 
First, we design six \textit{Difficulty-Enhancement Strategies} and train the \textbf{CoDiQ-Generator} via Reinforcement Learning to synthesize questions beyond zero-shot baselines. 
Second, we develop the \textit{CoDiQ Pipeline}, an iterative framework with hybrid verification to ensure logical consistency while increasing complexity. 
Third, we establish a relative difficulty paradigm through LLM-based ranking and a \textit{ValueNetwork} that quantifies difficulty via continuous scores for precise level grouping. Based on CoDiQ, we construct \textbf{CoDiQ-Corpus}, comprising 44K competition-grade math and coding question sequences. Human evaluation and experiments confirm that our method yields high-quality data that significantly enhances downstream reasoning performance.

Our key contributions are:

\begin{itemize}

\item \textbf{Difficulty-Enhancement Strategies}. We propose six systematic strategies that guide LLMs to inject difficult elements into question generation, enabling the synthesis of high-difficulty questions that surpass zero-shot generation baselines.

\item \textbf{Test-Time Scaling Tendency for Difficulty}. We identify a scaling tendency linking test-time compute to question difficulty, characterizing the upper bound of a model's capacity to produce valid, high-difficulty questions.

\item \textbf{CoDiQ-Corpus}. We construct a dataset of 44K competition-grade mathematical and coding questions based on our CoDiQ-Generator. Experiments demonstrate that training on CoDiQ-Corpus significantly enhances the reasoning capabilities of large reasoning models compared to existing baselines.

\end{itemize}

We will open-source CoDiQ-Corpus, CoDiQ-Generator, and all implementations to support future research.
\section{Related Works}

Generating difficult yet valid questions is increasingly recognized as a key lever for scaling reasoning progress: it expands the training distribution beyond scarce human-curated problems, continuously provides frontier-level supervision, and provides a controlled way of generalization testing under increasing difficulty~\cite{he2024olympiadbenchchallengingbenchmarkpromoting,sun2025challengingboundariesreasoningolympiadlevel,luong2025robustmathematicalreasoning}. As a result, recent research has devoted substantial effort to synthesizing competition-level problems with both intellectual challenge and formal correctness guarantees.

\paragraph{Prompt-based and agentic synthesis pipelines.}
One dominant paradigm treats hard-problem creation as a prompt-driven or agentic workflow: the system bootstraps from seed problems, concepts, or human-authored solution structures, then iteratively refines candidates with self-critique, filtering, and verification signals to ensure well-posedness~\cite{liu2024augmentingmathwordproblems,wang2025evolmathevalevolvablebenchmarksmathematical,zheng2025unicodeframeworkgeneratinghigh}. 
On the math side, PromptCoT~\cite{zhao2025promptcot} drives generation with concept sampling and structured design cues, explicitly inducing expert-like problem-construction rationales, and then applies rejection sampling to retain coherent, high-difficulty instances. CogAtom~\cite{chen2025cogatomcognitiveatomsolympiadlevel} instead decomposes human solutions into reusable cognitive atoms, constructs an atom graph, and synthesizes new problems via constrained recombination, enabling systematic exploration of a compositional design space.  
For programming tasks, reliability is even more salient: a valid instance requires not only a statement but also precise I/O specifications, meaningful constraints, and anti-shortcut test suites. AutoCode~\cite{zhou2025autocodellmsproblemsetters} exemplifies a closed-loop setter pipeline that jointly generates problem statements and reference solutions, and filters under-specified or ill-posed tasks via automated test generation and cross-verification.
Overall, these approaches are effective but often depend on complex multi-step orchestration and heavy post-hoc filtering to maintain validity.

\paragraph{Training  generators for difficult questions.}
A complementary line of work focuses on training dedicated generators to amortize the cost of multi-step agentic flow, enabling large-scale difficult-problem synthesis at low marginal cost~\cite{hu2025questincentivizingllmsgenerate,christ2024mathwellgeneratingeducationalmath,xie2024adversarialmathwordproblem}. 
For example, ScaleQuest~\cite{ding2025unleashingllmreasoningcapability} unlocks question-generation capability via Question Fine-Tuning and Question Preference Optimization to align generation toward solvability and difficulty. ScaleDiff~\cite{pei2025scalediffscalingdifficultproblems} first identifies hard instances efficiently , then trains a specialized generator  on the hard subset to expand the upper tail.
Overall, generator-training methods scale well, but common limitations remain: difficulty controls are often coarse, and validity still depends heavily on post-hoc filtering or human verification.

In contrast to prior synthesis pipelines and generator-training methods, our approach centers on test-time scaling as a core mechanism for fine-grained difficulty control under verifiable solvability: we explicitly scale instance difficulty at inference time while enforcing correctness via automated verification, rather than relying on filtering. 
This enables systematic frontier tracking of hard-yet-solvable questions while keeping validity constraints intact and controllable at scale.

\section{Method}
\subsection{Overview}
\begin{wrapfigure}{r}{0.45\textwidth}
    \centering
    \includegraphics[width=\linewidth]{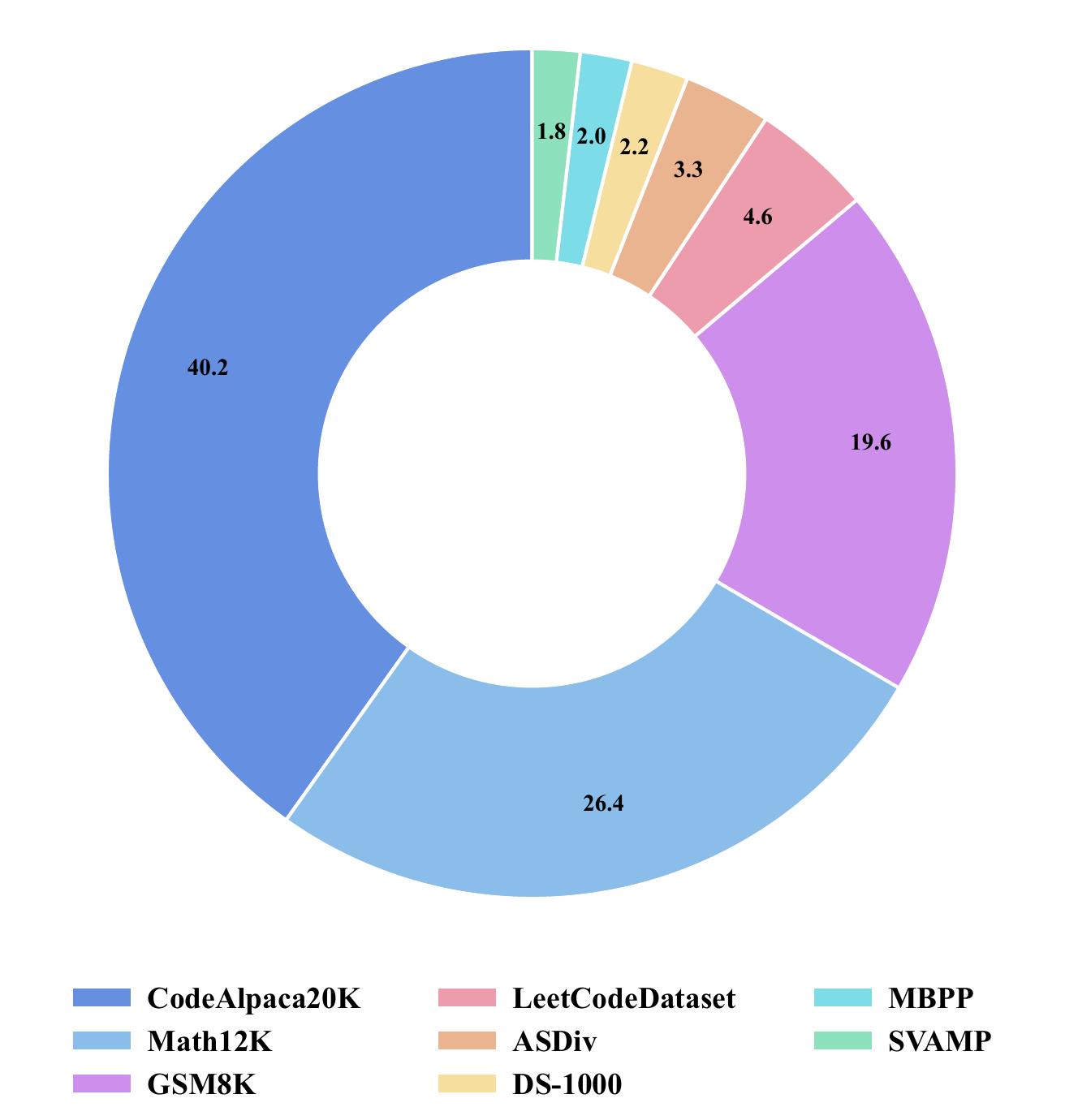}
    \caption{Distribution of CoDiQ-Corpus Dataset}
    \label{fig:dataset_distribution}
    \vspace{-0.2in}  % 减小这个值试试看
\end{wrapfigure}

Our method aims to endow LRMs with scalable test-time question generation capability by enabling them to synthesize progressively challenging yet valid questions. 
To achieve this, we first introduce six Difficulty-Enhancement Strategies (\S\ref{sec:strategies}) that explicitly guide LRMs toward difficulty-scaling reasoning and hard question construction. 
These strategies are instantiated within the CoDiQ Pipeline (\S\ref{sec:codiq_pipeline}), which integrates two verification modules—difficulty estimation (\S\ref{sec:difficulty_estimation}) and solvability verification (\S\ref{sec:solvability_verification})—to jointly regulate both difficulty and validity. 
Leveraging this pipeline, we construct CoDiQ-Bench to systematically benchmark models' question-generation performance under a unified evaluation framework. 
Then, we develop a specialized CoDiQ-Generator through reinforcement learning (\S\ref{sec:generator}), utilizing pipeline-derived feedback signals to further enhance the difficulty and reliability of synthesized questions. 
Finally, we construct CoDiQ-Corpus, a dataset of 44K competition-grade mathematical and coding questions based on our CoDiQ-Generator. The detailed statistics are provided in Appendix~\ref{appendix:codiq_corpus}, and the distribution is shown in Figure~\ref{fig:dataset_distribution}.

\subsection{Difficulty-Enhancement Strategies} 
\label{sec:strategies}
To systematically scale problem difficulty beyond naive prompting (e.g., ``make this harder''), we design six Difficulty-Enhancement Strategies (detailed in Appendix \ref{appendix:codiq_prompt}) that serve as explicit cognitive scaffolds for LLMs. These strategies—\textit{Dimensionality \& Constraints}, \textit{Mathematical Abstraction}, \textit{Inverse \& Constructive}, \textit{State Explosion}, \textit{Theorem Disguise}, and \textit{Edge Case \& Rigor Engineering}—guide the model to inject algorithmic difficulty systematically, ensuring difficulty arises from reasoning depth rather than superficial modifications.

\subsection{CoDiQ Pipeline}
\label{sec:codiq_pipeline}

Building upon the difficulty injection strategies (\S\ref{sec:strategies}), we introduce the CoDiQ Pipeline (Algorithm~\ref{alg:codiq_pipeline}), which systematically scales difficulty through iterative refinement. The pipeline implements an evolutionary loop where a seed question $Q_0$ progressively evolves into harder variants $\{Q_1, \dots, Q_n\}$ over up to $T_{\max} = 8$ rounds. At each iteration, the model is prompted with ``\textit{Can you make it more difficult?}'' to trigger deeper reasoning.

To ensure generation quality, the pipeline incorporates two core validation modules: \textit{Difficulty Estimation} (\S\ref{sec:difficulty_estimation}) and \textit{Solvability Verification} (\S\ref{sec:solvability_verification}). The process terminates under strict stopping rules (\S\ref{sec:termination_criteria}).

% CoDiQ Pipeline: 1 Difficulty Estimation.
\subsubsection{Difficulty Estimation}
\label{sec:difficulty_estimation}
\begin{wrapfigure}{r}{0.5\textwidth}
\begin{minipage}{0.5\textwidth}
\begin{algorithm}[H] % 注意这里必须是 [H]，不能是 [tb]
\caption{CoDiQ Iterative Pipeline}
\label{alg:codiq_pipeline}
\begin{algorithmic}[1]
   \STATE {\bfseries Input:} Seed Question $Q_0$, Max Rounds $T_{max}$
   \STATE {\bfseries Output:} Evolved Questions $\mathcal{Q}$

   \STATE Initialize $\mathcal{Q} \leftarrow \emptyset$, $d_0 \leftarrow \textsc{Difficulty}(Q_0)$

   \FOR{$i=1$ {\bfseries to} $T_{max}$}
      \STATE $Q_i \leftarrow \text{LLM}(Q_{i-1})$
      \STATE $d_i \leftarrow \textsc{Difficulty}(Q_i)$
      \IF{$\textsc{Valid}(Q_i) = \text{False}$ {\bfseries or} $d_i < d_{i-1}$}
         \STATE \textbf{break}
      \ENDIF
      \STATE $\mathcal{Q} \leftarrow \mathcal{Q} \cup \{Q_i\}$
   \ENDFOR

   \STATE \textbf{return} $\mathcal{Q}$
\end{algorithmic}
\end{algorithm}
\end{minipage}
\end{wrapfigure}
To strictly enforce the monotonic difficulty constraint, we require a robust mechanism to detect difficulty regression. Since CoDiQ targets the frontier of model capabilities, standard absolute scoring suffers from saturation effects---where models assign uniformly high scores to challenging queries---rendering direct comparison ineffective. Consequently, we adopt a \textit{relative} difficulty paradigm comprising two complementary approaches: explicit \textit{LLMs-Ranking} (\S\ref{sec:llms_rank}) to discern comparative hardness, and implicit \textit{ValueNetwork Scoring} (\S\ref{sec:valuenetwork}) to capture internal uncertainty. Finally, we normalize these discrete rankings (\S\ref{sec:normalization}) to eliminate granularity bias.

\paragraph{LLMs-Ranking.}
\label{sec:llms_rank}
To adaptively allocate the reasoning budget, we utilize Doubao-Seed-1.8~\cite{seedseed1} for listwise difficulty estimation. Given a batch of queries $\mathcal{Q} = \{q_1, \dots, q_n\}$, the model ranks them by perceived difficulty following the prompt template in Appendix~\ref{appendix:llms_ranking}. To mitigate positional bias, we apply \textit{stochastic shuffling} $\tau(\mathcal{Q})$ before ranking. The model outputs structured \texttt{JSON} results, from which we extract ranked indices to map computation budgets $K$, allocating more samples to harder queries.

\paragraph{ValueNetwork Scoring.}
\label{sec:valuenetwork}
To efficiently estimate question difficulty, we extend the hidden-representation-based approach of~\cite{zhu2025llm} by analyzing the model’s reasoning trajectory. We employ \textsc{Qwen3-8B} to extract hidden states across a sampling window of up to 4,096 tokens. To capture the critical early stages of reasoning, we implement a quadratic sampling strategy(Eq.\S\ref{eq:sampling}) that allocates higher density to the onset of generation. These representations are fed into a lightweight MLP trained via binary cross-entropy to predict the probability of correctness $y \in \{0, 1\}$ across a mixture of standard and competition-level benchmarks~\cite{cobbe2021gsm8k, numina_math_datasets, jain2024livecodebench, li2022competition}. This approach demonstrates a strong capability in distinguishing problem difficulty(See Appendix\ref{appendix:valuenetwork_performance}). At inference, the predicted probability serves as a proxy for LLM-perceived difficulty, where lower scores indicate higher difficulty. Detailed implementation is provided in Appendix\ref{appendix:training}.

\paragraph{Difficulty Normalization.}
\label{sec:normalization}
To convert the discrete grouped rankings from \S\ref{sec:llms_rank} and \S\ref{sec:valuenetwork} into continuous scores, we apply linear scaling. Given $G$ difficulty groups $\mathcal{G} = \{g_1, \dots, g_G\}$ ordered from easiest to hardest, the normalized difficulty for question $q_i$ in group $g_j$ is:
\begin{equation}
\label{eq:difficulty_computation}
d_i = \frac{j - 1}{G - 1}, \quad j \in \{1, \dots, G\}.
\end{equation}
This maps discrete rankings to $[0, 1]$, where $d_i$ serves as the scaling factor for adaptive computation allocation.

% CoDiQ Pipeline: 2 Solvability Verification.
\subsubsection{Solvability Verification}
\label{sec:solvability_verification}
While difficulty estimation ensures monotonic complexity growth, it does not guarantee logical validity. To prevent invalid or unsolvable instances, we utilize Qwen3-32B~\cite{yang2025qwen3} to verify the solvability of generated instances. Following the template in Appendix~\ref{appendix:solvability_check}, the model generates responses in \texttt{JSON} format, from which we extract the solvability status and confidence score. Only instances verified as solvable with high confidence are retained.

% CoDiQ Pipeline: 3 Termination Criteria.
\subsubsection{Termination Criteria}
\label{sec:termination_criteria}
To maintain the integrity of the question trajectory, the pipeline enforces strict stopping rules. The iterative process terminates immediately if: (1) \textbf{Non-Monotonic Difficulty}, where the generated question $Q_i$ has a lower difficulty score compared to its predecessors; or (2) \textbf{Unsolvability}, where the candidate $Q_i$ is flagged as invalid. Upon termination at step $i$, the invalid candidate is discarded, and the pipeline yields the sequence $\{Q_1, \dots, Q_{i-1}\}$. See Appendix~\ref{appendix:case_study} for case study and Appendix~\ref{appendix:failure_analysis} for failure type analysis.

% CoDiQ-Bench
\subsection{CoDiQ-Bench}
\label{sec:codiq_bench}
\begin{table}[htbp]
\centering
\caption{Dataset statistics of CoDiQ-Bench.}
\begin{tabular}{cc}
\toprule
\textbf{Statistics} & \textbf{Number} \\
\midrule
\textbf{\#Questions} & $200$ \\
~~~~- \textit{math}   & $100$\\
~~~~- \textit{code}   & $100$\\
\midrule
\textbf{Question Tokens Length} \\
~~~~- \textit{max/min/avg}   & $726$/$9$/$128.2$ \\
\bottomrule
\end{tabular}
\label{tab:codiqbench_statistics}
\end{table}
To systematically evaluate the question generation capability of LRMs, we first construct CoDiQ-Bench, a curated dataset comprising 200 carefully selected cases across coding and mathematical domains (Table\S\ref{tab:codiqbench_statistics}). For coding tasks, we randomly sample 50 cases each from CodeAlpaca\_20K (general programming) and LeetCodeDataset (algorithmic challenges). For mathematical reasoning, we sample 50 cases each from GSM8K (grade school questions) and MATH12K (mathematical question-solving). We intentionally focus on relatively simple questions to establish a baseline benchmark, with detailed selection criteria regarding solvability and quality provided in Appendix\S\ref{appendix:bench_selection}.

\subsection{CoDiQ-Generator}
\label{sec:generator}
To further enhance the CoDiQ Pipeline's capacity for generating high-difficulty, high-quality questions, we develop CoDiQ-Generator via reinforcement learning. By directly optimizing the model's question-setting behavior through targeted reward signals, we aim to improve both the validity and difficulty scaling of synthesized problems.

\subsubsection{RL Data Construction}
\label{sec:rl_data}

We construct our Reinforcement Learning dataset, $\mathcal{D}_{RL}$, by capturing the critical failure modes of Qwen3-8B within the CoDiQ Pipeline (Section \ref{sec:codiq_pipeline}). Rather than maximizing absolute difficulty, we target the model's specific \textit{capability boundary}~\cite{zhang2025interplay}. We identify evolutionary trajectories where the model successfully generates valid questions for rounds $1$ through $i-1$ but fails at round $i$ (e.g., due to unsolvability or difficulty stagnation). These boundary instances are collected to form training pairs, effectively converting the model's ``breaking point'' into a precise learning signal.

To ensure broad domain coverage, we initialize the pipeline with seed questions ($Q_0$) drawn from diverse established benchmarks. For mathematics, we sample from \textsc{Math12k}~\cite{hiyouga_math12k}, \textsc{GSM8K}~\cite{cobbe2021gsm8k}, \textsc{SVAMP}~\cite{ChilleD_SVAMP}, and \textsc{ASDiv}~\cite{yimingzhang_asdiv}. For code generation, we utilize \textsc{Code Alpaca}~\cite{codealpaca}, \textsc{LeetCodeDataset}~\cite{xia2025leetcodedataset}, \textsc{MBPP}~\cite{austin2021program}, and \textsc{DS-1000}~\cite{Lai2022DS1000}. After filtering for the specific boundary conditions described above, the final dataset $\mathcal{D}_{RL}$ comprises \textbf{1,173} high-quality samples.

\subsubsection{RL Training Paradigm}
\paragraph{Reinforcement Learning Optimization (RL).} 
The recent success of R1-style methods has demonstrated the effectiveness of online RL using discrete, rule-based rewards~\cite{shao2024deepseekmath}. In our pipeline, Qwen3-8B~\cite{yang2025qwen3} is further refined using reinforcement learning signals derived from solvability confidence, difficulty progression, and question validity checks. Based on the dataset described in \S\ref{sec:rl_data}, we apply a rule-based RL approach to optimize the model's judgment reasoning capability. Specifically, we utilize the GRPO algorithm~\cite{shao2024deepseekmath} within the VeRL reinforcement learning framework~\cite{sheng2025hybridflow}.

To ensure smooth optimization, we design a difficulty-aware reward function that balances validity guarantees with progressive difficulty scaling. Given confidence lower bound 
$$\text{conf} = \max(0.5, \text{confidence}(x))$$ 
and difficulty change 
\begin{equation}
\Delta(D) = d_i - d_{i-1} \in [-1, 1]
\end{equation}

for iteration $i \in \{1, \dots, R\}$, where $d_i$ is computed via Eq.~\eqref{eq:difficulty_computation} and $R$ denotes the maximum number of evolution rounds:

\vspace{-4mm}
\begin{small}
\begin{equation}
r = \begin{cases}
    0, & \text{if invalid}\\
    0.6 \cdot \text{conf}, & \text{if } \Delta(D) = 0\\
    0.2 \cdot \text{conf} + 0.8 \cdot (0.8 + 0.2 \cdot \Delta(D)), & \text{if } \Delta(D) > 0
\end{cases}
\label{eq:reward_function}
\end{equation}
\end{small}
\vspace{-4mm}

\noindent
where invalid cases include unsolvable questions, repetitive outputs, or negative difficulty changes ($\Delta(D) < 0$).

\section{Experiments}
\subsection{Experimental Setup}
\paragraph{Baselines} To evaluate the effectiveness of our CoDiQ Prompt and CoDiQ-Generator, we compare against models with inherent test-time scaling capabilities that support extended reasoning. These baseline models include flagship models (GLM-4.6~\cite{glm2024chatglm}) and smaller-parameter models (GPT-OSS-20B~\cite{agarwal2025gpt}, GLM-Z1-9B-0414~\cite{glm2024chatglm}, and the Qwen3 series~\cite{yang2025qwen3}: Qwen3-0.4B, Qwen3-1.7B, Qwen3-4B, Qwen3-8B, Qwen3-14B, Qwen3-32B). All these models utilize the CoDiQ Pipeline described in Section~\ref{sec:codiq_pipeline} for generation.

\paragraph{Evaluation Metrics.}
\label{sec:evaluation_metrics}
We employ two metrics to quantify problem difficulty: 
(1) \textbf{DS-LLM}: Difficulty score estimated by the Doubao-Seed-1.8~\cite{seedseed1} model (details in Section~\ref{sec:llms_rank}). 
(2) \textbf{DS-VN}: Difficulty score derived from the ValueNetwork (VN) (details in Section~\ref{sec:valuenetwork}). 
Both scores are normalized to the range [0, 1] and reported as percentages (0\%-100\%), where higher values indicate greater difficulty. All reported scores are averaged across questions in CoDiQ-Bench.

\subsection{Main Results}
\subsubsection{Maximum Solvable Difficulty}

\begin{table}[!tb]
\centering
\caption{\textbf{Performance of different Long-CoT models on CoDiQ-Bench.}
Group rankings based on the highest difficulty of solvable questions generated across 8 rounds without difficulty degradation on CoDiQ-Bench.
The best, the second-best and the third-best scores for each indicator are shown in  \boxed{box}, \textbf{bold} and \underline{underlined}, respectively.}
\begin{adjustbox}{width=0.8\linewidth, center}
\begin{tabular}{cccccc}
\toprule
Model & Rounds & Tokens & DR-LLM & DR-VN & DR(AVG) \\
\midrule
\multicolumn{6}{c}{\textbf{\textit{Direct Prompt}}} \\
\midrule
\rowcolor{color12}
GPT-OSS-20B & 2.9 & 5528.2 & 68.5 & \textbf{74.4} & \textbf{71.5} \\
\rowcolor{color12}
GLM-4.6 & 2.8 & 3385.8 & \textbf{71.2} & \underline{65.8} & \underline{68.5} \\
\rowcolor{color12}
Qwen3-32B & 2.3 & 1239.3 & 50.6 & 54.8 & 52.7 \\
\rowcolor{color12}
Qwen3-8B & \underline{3.4} & 1130.5 & 39.2 & 59.6 & 49.4 \\
\rowcolor{color12}

GLM-Z1-9B-0414 & 2.7 & 1229.8 & 48.8 & 43.7 & 46.3 \\
\rowcolor{color12}
Qwen3-14B & 3.1 & 2076.4 & 45.9 & 44.4 & 45.2 \\
\rowcolor{color12}
Qwen3-4B & \boxed{4.2} & 1419.7 & 36.8 & 40.4 & 38.6 \\
\rowcolor{color12}
Qwen3-1.7B & 3.3 & 844.5 & 25.6 & 37.1 & 31.4 \\
\rowcolor{color12}
Qwen3-0.6B & 2.4 & 314.3 & 17.2 & 35.0 & 26.1 \\

\midrule
\multicolumn{6}{c}{\textbf{\textit{CoDiQ Prompt(ours)}}} \\
\midrule
% \rowcolor{color22}
% DeepSeek-V3.2 & 1.7 & 5747.2 & \underline{68.8} & 63.5 & 66.2 \\
\rowcolor{color22}
GLM-4.6 & 2.7 & \underline{7143.8} & \boxed{73.2} & \boxed{83.3} & \boxed{78.3} \\
\rowcolor{color22}
GPT-OSS-20B & 2.1 & \boxed{8057.3} & 63.8 & 61.5 & 62.7 \\
\rowcolor{color22}
Qwen3-32B & 2.2 & 4893.6 & 63.0 & 46.5 & 54.8 \\
\rowcolor{color22}
Qwen3-14B & 2.6 & 5281.9 & 53.9 & 44.2 & 49.1 \\
\rowcolor{color22}
Qwen3-4B & 2.8 & 4422.3 & 49.1 & 42.7 & 45.9 \\
\rowcolor{color22}
Qwen3-8B & 2.4 & 4155.6 & 49.8 & 41.9 & 45.8 \\
\rowcolor{color22}
GLM-Z1-9B-0414 & 1.7 & 3638.3 & 54.7 & 30.0 & 42.4 \\
\rowcolor{color22}
Qwen3-1.7B & 1.4 & 2975.7 & 32.3 & 37.3 & 34.8 \\
\rowcolor{color22}
Qwen3-0.6B & 1.0 & 2052.7 & 22.4 & 29.2 & 25.8 \\

\midrule
\multicolumn{6}{c}{\textbf{\textit{CoDiQ Generator(ours)}}} \\
\midrule
\rowcolor{color32}
CoDiQ-Gen-8B & \textbf{3.4} & \textbf{7499.6} & 58.9 & 58.1 & 58.5 \\
\bottomrule
\end{tabular}
\end{adjustbox}
\label{tab:main}
\end{table}
% CoDiQ-Generator 同样实验设置下，生成的题目难度更难
\label{sec:max_difficulty}
To evaluate the question generation capability of Large Reasoning Models (LRMs) within our proposed framework, and to identify the optimal Generator for the subsequent synthesis of difficult questions, we conduct a comparative analysis. Specifically, we instantiate distinct LRMs as the backbone Generator within the CoDiQ Pipeline and assess the difficulty of the questions they generate on CoDiQ-Bench.

\paragraph{Effectiveness of CoDiQ Prompt.} We first evaluate the efficacy of the CoDiQ Prompt in eliciting deep reasoning for difficulty synthesis. As detailed in Table~\ref{tab:main}, the application of the CoDiQ Prompt induces a substantial expansion in reasoning token usage across all evaluated architectures. This significant increase in test-time computation suggests that the prompt successfully triggers extended reasoning trajectories, enabling models to construct more intricate constraints and logic. Consequently, the majority of baseline models exhibit a marked improvement in the difficulty of generated questions when conditioned on our prompt.

\paragraph{Superiority of CoDiQ-Generator.} Notably, our CoDiQ-Gen-8B outperforms the significantly larger Qwen3-32B in generating high-complexity instances. We attribute this performance gain to the Reinforcement Learning alignment described in Section~\ref{sec:rl_data}. By optimizing for solvability and difficulty progression, the RL training enables CoDiQ-Generator to maintain high validity rates across iterative evolution. This stability allows the model to sustain the generation pipeline for a greater number of rounds—exceeding the iteration depth of baseline models—thereby accumulating complexity monotonically without premature termination due to unsolvability.

\subsubsection{Difficulty Metrics Comparison}
\label{sec:diff_metric}
\begin{figure}[t]
    \centering
    \begin{minipage}{0.48\textwidth}
        \centering
        \includegraphics[width=\linewidth]{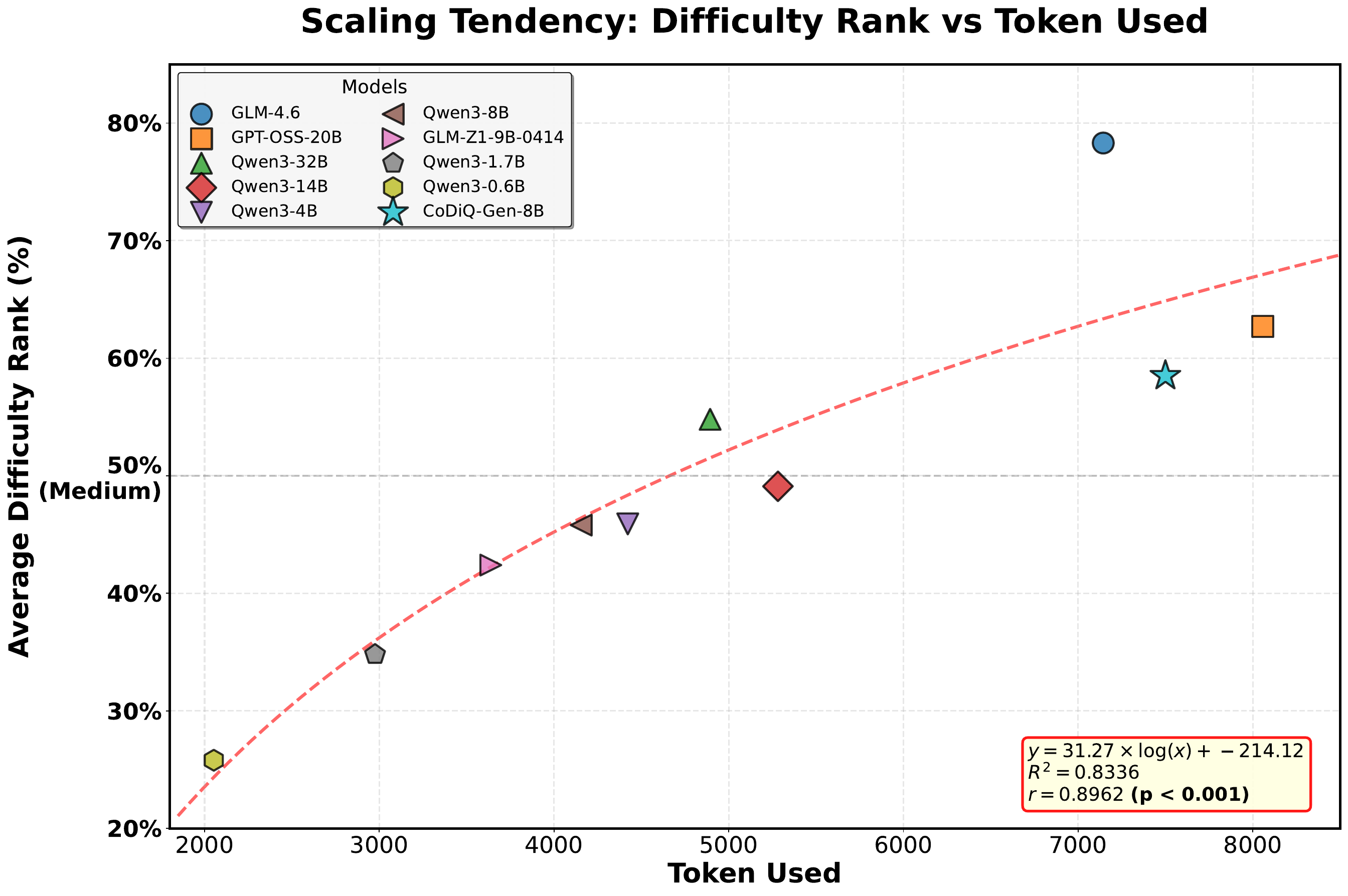}
        \caption{\textbf{Question Difficulty Scaling on CoDiQ-Bench.} 
        Scatter plot showing the relationship between average reasoning tokens and difficulty ranking (DR-AVG) for models using CoDiQ Prompt. Each point represents a model, demonstrating the positive correlation between increased reasoning computation and generated problem difficulty.}
        \label{fig:scaling_tendency_discrete_fitting}
    \end{minipage}
    \hfill
    \begin{minipage}{0.48\textwidth}
        \centering
        \includegraphics[width=\linewidth]{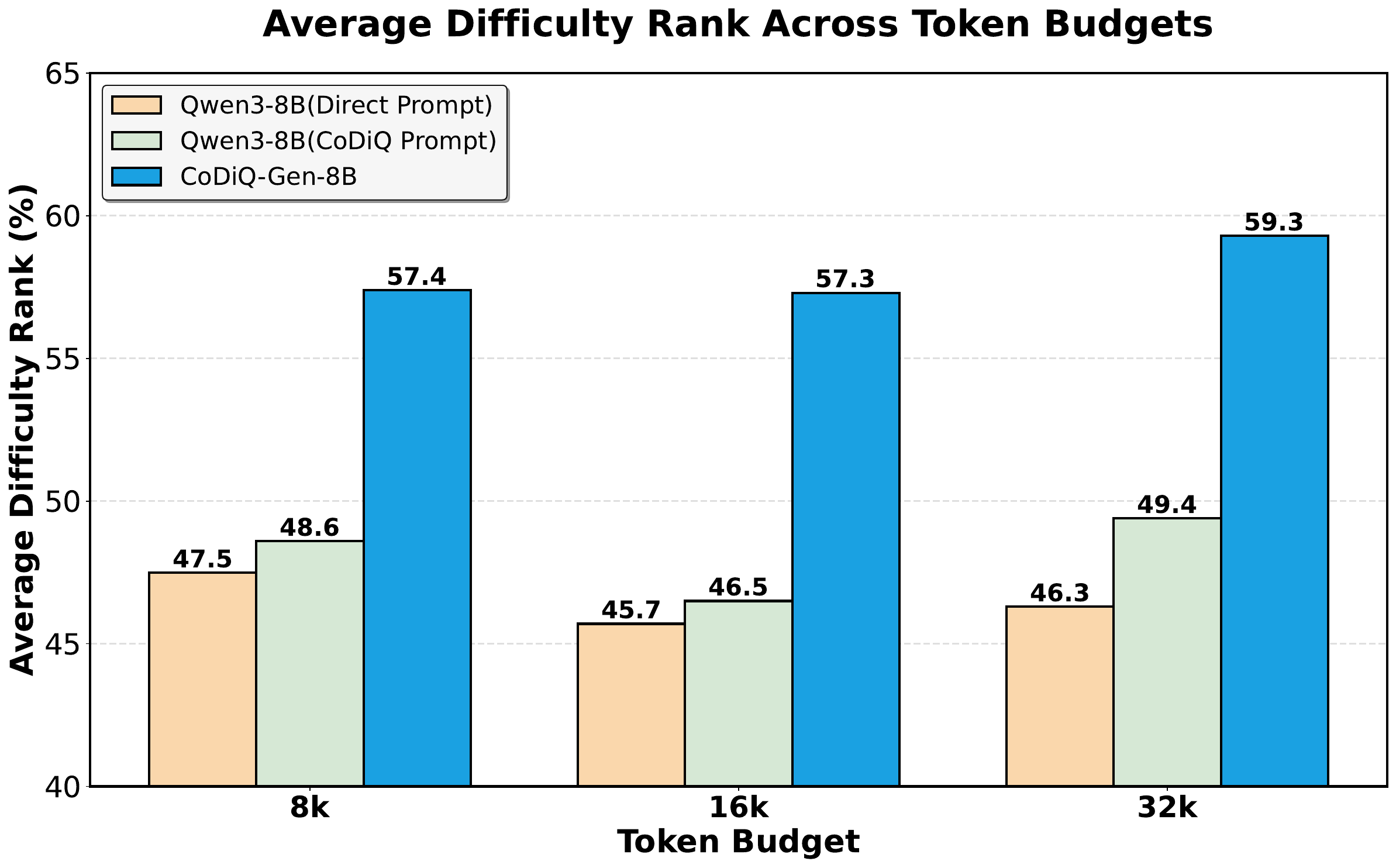}
        \caption{\textbf{Model Performance on CoDiQ-Bench Across Token Budgets.} 
        Average difficulty rank (\%) of three model variants (Qwen3-8B with Direct Prompt, 
        Qwen3-8B with CoDiQ Prompt, and CoDiQ-Gen-8B) under different token budget constraints 
        (8k, 16k, 32k). Higher scores indicate better performance in handling difficult questions.}
        \label{fig:model_performance_comparison}
    \end{minipage}
\end{figure}
To further verify the number of tokens consumed by LRMs can estimate question difficulty, we highlight the positive correlation between token volume and difficulty rankings shown in Figure~\ref{fig:scaling_tendency_discrete_fitting}. We further validated this relationship by analyzing the correlation between token consumption and our established metrics (DR-LLM and DR-VN), yielding Pearson coefficients of $r = 0.8299$ ($p < 0.001$) and $r = 0.8545$ ($p < 0.001$), respectively. These results confirm that computational cost serves as a reliable proxy for difficulty, provided that the problem complexity remains within the evaluator's capability and a consistent scaling method is applied.

% \subsubsection{Difficulty Metrics Comparison}
% \label{sec:diff_metric}
% To further verify the number of tokens consumed by LRMs can estimate question difficulty, we calculated the Pearson correlation between token consumption and our difficulty metrics (DR-LLM and DR-VN), which are 0.8299 and 0.8545. These statistically significant results validate the scientific nature of our method, confirming that computational cost is a reliable proxy for difficulty in this context.

% \subsubsection{Difficulty Metrics Comparison}
% To assess the consistency of our difficulty metrics, we compute the Pearson correlation coefficient between DR-LLM and DR-VN based on model performance across CoDiQ-Bench. The coefficient of $r = 0.7481$ ($p = 0.00023$) demonstrates strong agreement between the two metrics, confirming that both approaches reliably capture problem difficulty. 

\subsection{Ablation Study}
\subsubsection{Upper Bound of Difficulty Generation}
\label{sec:upper_bound}
\begin{table}[!tb]
\centering
\caption{\textbf{Performance of different Long-CoT models on CoDiQ-Bench.}
Group rankings based on the highest difficulty of questions generated across 8 rounds on CoDiQ-Bench.
The best, the second-best and the third-best scores for each indicator are shown in  \boxed{box}, \textbf{bold} and \underline{underlined}, respectively.}
\begin{adjustbox}{width=0.8\linewidth, center}
\begin{tabular}{cccccc}
\toprule
Model & Rounds & Tokens & DR-LLM & DR-VN & DR(AVG) \\
\midrule
\multicolumn{6}{c}{\textbf{\textit{Direct Prompt}}} \\
\midrule
\rowcolor{color12}
Qwen3-8B & \boxed{6.0} & 2439.7 & 33.5 & 39.1 & 36.3 \\
\rowcolor{color12}
Qwen3-14B & 5.6 & 4927.4 & 45.6 & 55.6 & 50.6 \\
\rowcolor{color12}
Qwen3-32B & 5.7 & 4124.9 & \textbf{65.3} & 47.5 & 56.4 \\

\midrule
\multicolumn{6}{c}{\textbf{\textit{CoDiQ Prompt(ours)}}} \\
\midrule
\rowcolor{color22}
Qwen3-8B & \underline{5.8} & 7282.2 & 53.5 & 53.3 & 53.4 \\
\rowcolor{color22}
Qwen3-14B & 5.6 & \underline{9590.2} & \underline{58.6} & \underline{63.1} & \underline{60.9} \\
\rowcolor{color22}
Qwen3-32B & 5.7 & \textbf{9762.4} & \boxed{74.6} & \textbf{65.0} & \boxed{69.8} \\
\midrule
\multicolumn{6}{c}{\textbf{\textit{CoDiQ Generator(ours)}}} \\
\midrule
\rowcolor{color32}
CoDiQ-Gen-8B & \textbf{5.9} & \boxed{12591.6} & 52.6 & \boxed{72.2} & \textbf{62.4} \\
\bottomrule
\end{tabular}
\end{adjustbox}
\label{tab:generation_strategy}
\end{table}

In \S~\ref{sec:max_difficulty}, we evaluated the maximum solvable difficulty under the constraint of maintaining solution validity. However, this solvability requirement inherently limits the difficulty ceiling, as highly complex questions may not be unsolvable per se, but rather beyond the current model's capability to generate valid solutions. To explore the theoretical upper bound of difficulty synthesis—independent of solution generation constraints—we conduct an ablation study by removing the solvability verification module from the CoDiQ Pipeline. 

The results indicate that incorporating the CoDiQ Prompt significantly elevates the difficulty ceiling across backbone models compared to standard prompting. Notably, despite having fewer parameters, our CoDiQ-Gen-8B generates questions with a difficulty upper bound that surpasses that of Qwen3-14B. This suggests that our specialized tuning and prompting strategy effectively unlocks the potential for synthesizing highly complex logical structures, even in smaller architectures.

% Token Budget
\subsubsection{Impact of Max Token Budget}

We further examine the efficiency of difficulty scaling relative to computational cost. Figure~\ref{fig:model_performance_comparison} illustrates the maximum difficulty of \textit{solvable} questions generated by the CoDiQ Pipeline under strict constraints on accumulated token usage. 
To simulate resource-constrained environments, we enforce a strict cumulative token budget that encompasses both generation and verification phases. If the total token consumption exceeds the threshold during an iteration, that round is discarded, and the system reports the highest-difficulty valid problem from the preceding rounds.
The comparative analysis reveals that CoDiQ-Gen-8B exhibits a distinct advantage across all token budget thresholds, consistently yielding higher difficulty scores than baseline models. Furthermore, we observe that Qwen3-8B utilizing the CoDiQ Prompt achieves superior performance compared to its direct prompt counterpart. This performance gap validates the effectiveness of our CoDiQ methodology in leveraging computational resources to maximize question difficulty while maintaining solvability.

\subsection{Scaling Tendency Analysis}
\label{sec:scaling_tendency}
\begin{figure*}[htbp]
    \centering
    \begin{minipage}[t]{0.48\textwidth}
        \centering
        \includegraphics[width=\linewidth]{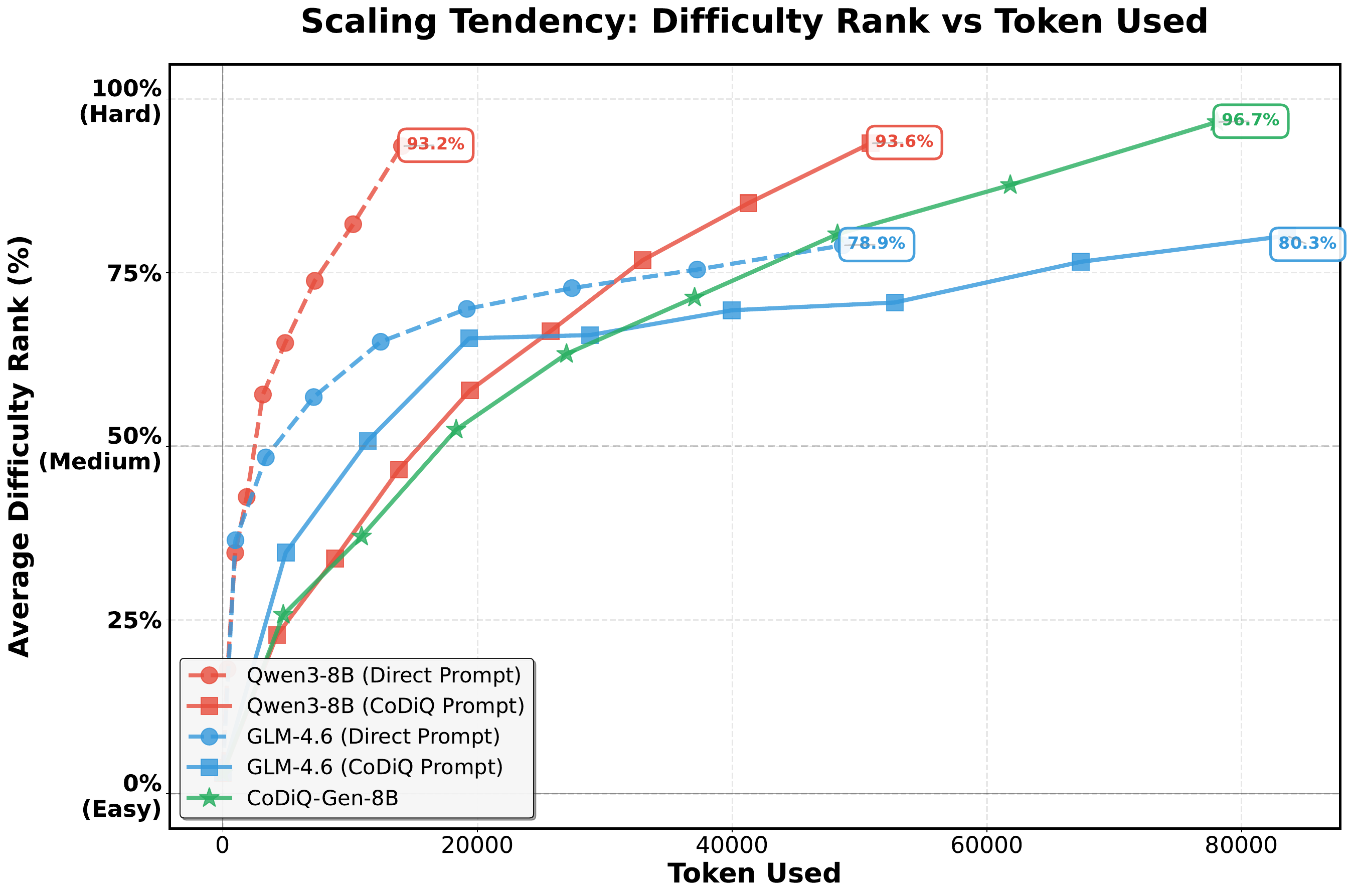}
        \caption{\textbf{Question Difficulty Scaling on CoDiQ-Bench.} Normalized average difficulty ranking of questions generated by different Long-CoT models across 8 rounds. Higher rankings indicate higher question difficulty and better model performance.}
        \label{fig:scaling_tendency_difficulty_combined}
    \end{minipage}
    \hfill
    \begin{minipage}[t]{0.48\textwidth}
        \centering
        \includegraphics[width=\linewidth]{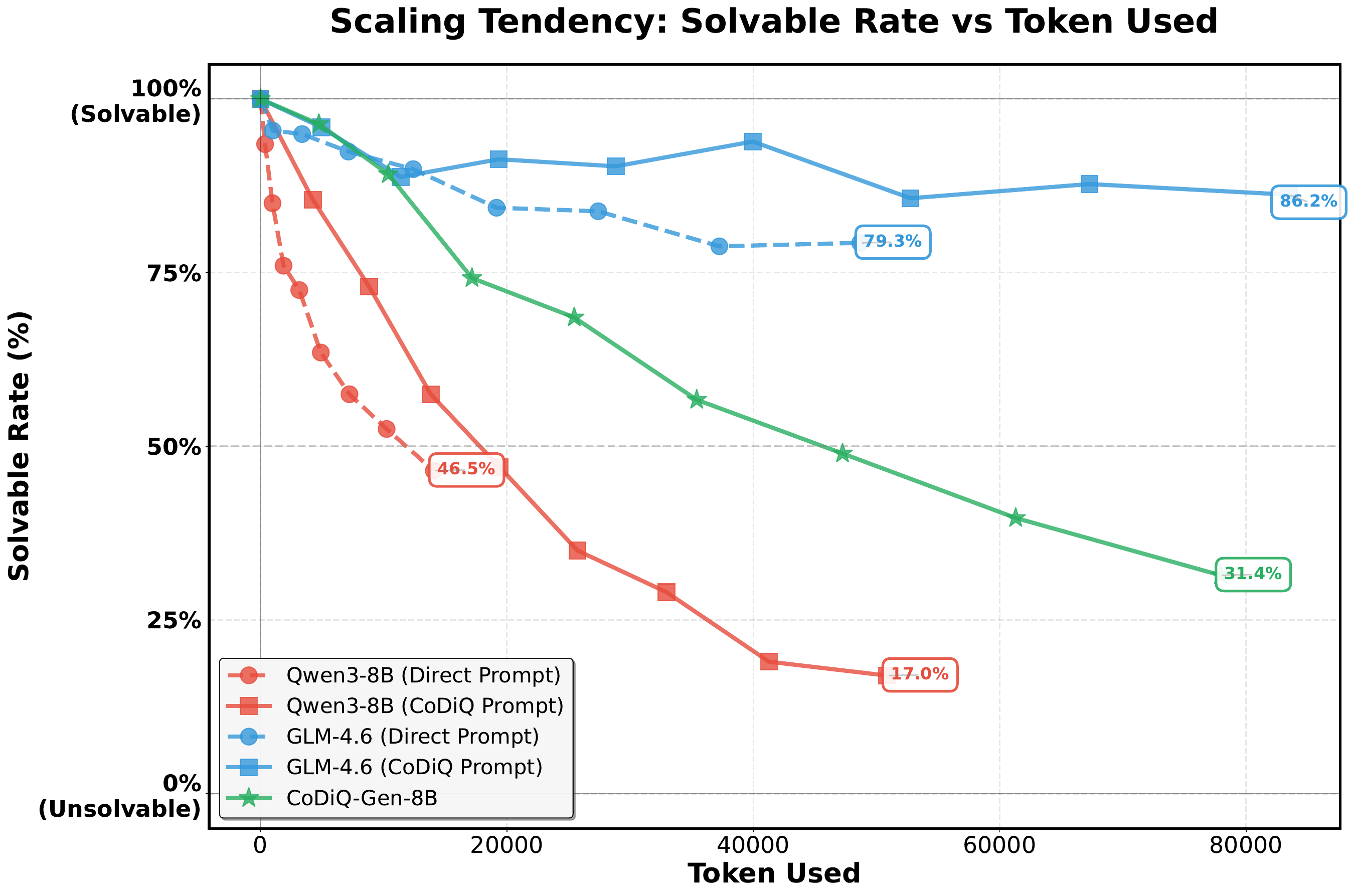}
        \caption{\textbf{Question Solvability Scaling on CoDiQ-Bench.} Solvable rate of questions generated by different Long-CoT models across 8 rounds. Higher indicates indicate better question quality.}
        \label{fig:scaling_tendency_solvable_combined}
    \end{minipage}
\end{figure*}

The preceding analyses established the performance ceilings of different LRMs, identifying both their maximum solvable difficulty (\S~\ref{sec:max_difficulty}) and their theoretical upper bounds (\S~\ref{sec:upper_bound}). However, these metrics represent static endpoints. To understand how these models arrive at such complexity, we now shift to a fine-grained analysis of the generation dynamics. In this section, we track the scaling tendencies of difficulty and solvability relative to reasoning computation within specific model groups (More results are provided in Appendix~\ref{appendix:scaling_tendency}).

% \subsubsection{Difficulty Scaling}
% \label{sec:diff_scaling}
% We analyze problem complexity evolution across 8 generation rounds (Figure~\ref{fig:scaling_tendency_difficulty_combined}) by mapping difficulty rankings against cumulative reasoning tokens. Under the \textit{Direct Prompt}, larger models saturate early, likely due to reaching the upper bound of either the model's generation capability or the difficulty evaluator after substantial token consumption. Conversely, the \textit{CoDiQ Prompt} stimulates extended reasoning, enabling smaller models to achieve higher complexity earlier. A unified scaling law emerges: increased reasoning computation consistently correlates with higher relative difficulty. Crucially, this observation corroborates the metric validation in Section~\ref{sec:diff_metric} from a single-model perspective, further confirming that token volume serves as a robust indicator of difficulty.

\subsubsection{Difficulty Scaling}
\label{sec:diff_scaling}
We analyze problem complexity evolution across 8 generation rounds in Figure~\ref{fig:scaling_tendency_difficulty_combined}. Compared to the \textit{Direct Prompt}, the \textit{CoDiQ Prompt} significantly stimulates deeper reasoning, resulting in a marked increase in token consumption. While a consistent upward difficulty trajectory is observed across most models, large-parameter models tend to saturate in later rounds. We attribute this plateau to the substantial token consumption, which likely approaches the upper bound of either the model's generation capacity or the difficulty evaluator's limit. Furthermore, this analysis corroborates the findings in Section~\ref{sec:diff_metric} from a single-model perspective, reinforcing the conclusion that token volume serves as a robust indicator of difficulty.

\subsubsection{Solvability Scaling}
We examine how solvability rates degrade with increasing difficulty (Figure~\ref{fig:scaling_tendency_solvable_combined}). This degradation reveals a fundamental trade-off between problem difficulty and validity. Three key findings emerge:

\begin{itemize}
    \item \textbf{Robustness of SOTA Models:} Flagship models (e.g., GLM-4.6) maintain high solvability across all difficulty levels, demonstrating well-balanced generation-verification capabilities.
    \item \textbf{Over-Reasoning Pitfall:} Smaller models experience validity collapse under CoDiQ, as they generate complexity beyond their reasoning capacity.
    \item \textbf{Efficacy of RL Alignment:} CoDiQ-Gen-8B breaks this degradation pattern through RL, successfully decoupling difficulty scaling from validity loss.
\end{itemize}

\subsection{Effectiveness of CoDiQ-Corpus}
To comprehensively assess the value of this corpus, we conduct a multi-dimensional evaluation focusing on \textit{difficulty}(Section\S\ref{sec:corpus_difficulty}), \textit{quality}(Section\S\ref{sec:corpus_quality}), and \textit{training effectiveness}(Section\S\ref{sec:corpus_effectiveness}). We first demonstrate that CoDiQ-Corpus significantly surpasses existing competition-grade benchmarks in problem hardness. Subsequently, we verify the logical soundness and solvability of the generated problems through rigorous human evaluation. Finally, we validate the practical utility of the corpus by employing it in a curriculum learning framework, demonstrating its capability to drive continuous improvements in reasoning models.

% Difficulty Comparison
\subsubsection{Difficulty Comparison}
\label{sec:corpus_difficulty}
To validate the elevated difficulty of CoDiQ-Corpus, we randomly sample 300 questions from each dataset, including CoDiQ-Corpus, AIME~\cite{aime_1983_2024}, NuminaMath-1.5~\cite{numina_math_datasets}, LiveCodeBench~\cite{jain2024livecodebench}, and Code-Contests~\cite{li2022competition}, and compare them using the ranking methodology in Section~\ref{sec:evaluation_metrics}. As shown in Table~\ref{tab:codiq_difficulty}, our CoDiQ-Corpus demonstrates significantly higher difficulty than existing competition-level datasets.
\begin{table}[htbp!]
\centering
\begin{minipage}[t]{0.48\textwidth}
\centering
\caption{\textbf{Datasets Difficulty Comparison.} The best, the second-best and the third-best scores for each indicator are shown in \boxed{box}, \textbf{bold} and \underline{underlined}, respectively.}
\label{tab:codiq_difficulty}
\begin{tabular}{lccc}
\toprule
Dataset & DR-LLM & DR-VN & DR(AVG) \\
\midrule
\multicolumn{4}{c}{\textbf{\textit{Baselines}}} \\
\midrule
\rowcolor{color12}
AIME(1983-2024) & \textbf{57.9} & \underline{45.1} & \textbf{51.5} \\
\rowcolor{color12}
NuminaMath-1.5 & 27.5 & 32.0 & 29.8 \\
\rowcolor{color12}
LiveCodeBench & 39.4 & \textbf{45.2} & 42.3 \\
\rowcolor{color12}
Code-Contests & \underline{47.2} & 41.0 & 44.1 \\
\midrule
\multicolumn{4}{c}{\textbf{\textit{CoDiQ Dataset(ours)}}} \\
\midrule
\rowcolor{color32}
CoDiQ-Corpus & \boxed{91.4} & \boxed{82.8} & \boxed{87.1} \\
\bottomrule
\end{tabular}
\end{minipage}
\hfill
\begin{minipage}[t]{0.48\textwidth}
\centering
\caption{\textbf{Model Performance Comparison.} The best, the second-best and the third-best scores for each indicator are shown in \boxed{box}, \textbf{bold} and \underline{underlined}, respectively.}
\label{tab:rl_model_comparison}
\begin{tabular}{lcc}
\toprule
\textbf{Model} & \textbf{MATH-500} & \textbf{AIME 2024} \\
\midrule
\multicolumn{3}{c}{\textbf{\textit{Baselines}}} \\
\midrule
\rowcolor{color12}
Qwen3-4B & 94.4 & 63.1 \\
\rowcolor{color12}
Qwen3-RL-4B & \underline{95.2} & 64.3 \\
\midrule
\multicolumn{3}{c}{\textbf{\textit{Curriculum Learning Models(ours)}}} \\
\midrule
\rowcolor{color32}
CoDiQ-L1-4B & \textbf{96.0} & \underline{65.0} \\
\rowcolor{color32}
CoDiQ-L2-4B & 94.8 & \textbf{66.7} \\
\rowcolor{color32}
CoDiQ-L3-4B & \boxed{96.0} & \boxed{70.6} \\
\bottomrule
\end{tabular}
\end{minipage}
\end{table}

% Human Quality Assessment
\subsubsection{Human Quality Assessment}
\label{sec:corpus_quality}
To verify the reliability of our CoDiQ-Corpus and CoDiQ Pipeline, we conducted human evaluation on $N=200$ stratified samples from accepted CoDiQ-Corpus and rejected cases. Three PhD experts independently assessed Clarity, Completeness, and Reasoning Validity (Appendix~\ref{appendix:data_quality_criteria}), achieving substantial agreement (Fleiss' $\kappa = 0.76$). 

Results show 82\% precision for accepted instances and 90\% NPV for rejected cases. Notably, error analysis on the false negatives (valid problems incorrectly rejected) empirically reveals the \textit{Verifier Paradox}: these instances were logically sound but exceeded the verifier's reasoning horizon, causing the model to misclassify them as ``unsolvable'' rather than ``hard.'' This confirms that our pipeline's upper bound is currently capped by the verifier's capability.

\subsubsection{Training Effectiveness Validation}
\label{sec:corpus_effectiveness}

% \paragraph{Reinforcement Learning Validation via Curriculum.}
% A distinct advantage of CoDiQ is its inherent controllability. By adjusting the token budget, it generates question sequences of progressive difficulty. This naturally facilitates a curriculum learning strategy~\cite{bengio2009curriculum} aligned with the model's evolving capabilities.

% Leveraging this difficulty-stratified data, we implement a multi-stage reinforcement learning paradigm. We sequentially train models, denoted as CoDiQ-L$_i$-4B ($i \in \{1,2,3\}$), where each stage $i$ incorporates a dataset subset of progressively higher difficulty. Reward signals are derived by prompting Qwen3-32B to evaluate response quality across multiple dimensions with weighted aggregation. Detailed training configurations are provided in Appendix~\ref{appendix:curriculum_learning_detail}. To validate the efficacy of this approach, we compare our model against two baselines: the vanilla Qwen-4B and Qwen3-RL-4B, a model trained via standard reinforcement learning on the original (non-transformed) datasets without difficulty stratification. Evaluation results on the MATH-500 and AIME 2024 benchmarks, presented in Table~\ref{tab:rl_model_comparison}, demonstrate that our budget-controlled curriculum learning framework significantly enhances model performance compared to standard training paradigms.

\paragraph{Reinforcement Learning Validation via Curriculum.}
A distinct advantage of CoDiQ lies in its inherent controllability. By adjusting the token budget, it generates question sequences of progressive difficulty, naturally facilitating a curriculum learning strategy~\cite{bengio2009curriculum} that aligns with the model's evolving capabilities. 

Leveraging this, we implement a multi-stage reinforcement learning paradigm by sequentially training models CoDiQ-L$_i$-4B ($i \in \{1,2,3\}$), where each stage $i$ utilizes a dataset subset of increasing difficulty. Rewards are derived by prompting Qwen3-32B to evaluate response quality via weighted aggregation (details in Appendix~\ref{appendix:curriculum_learning_detail}). We compare our approach against vanilla Qwen-4B and Qwen3-RL-4B, a baseline trained via standard RL on original datasets without stratification. Evaluation results on MATH-500 and AIME 2024 (Table~\ref{tab:rl_model_comparison}) demonstrate that our budget-controlled curriculum framework significantly enhances performance compared to standard training paradigms, thereby validating the effectiveness and utility of our CoDiQ-Corpus.

\section{Conclusion \& Limitations}
We presented \textit{CoDiQ}, a principled framework for synthesizing verifiable, high-difficulty reasoning problems at scale. By addressing the generator capacity ceiling through test-time scaling and mitigating "fake hard" instances via a hybrid verification pipeline, we successfully trained the \textbf{CoDiQ-Generator} using reinforcement learning. The resulting \textbf{CoDiQ-Corpus} features budget-driven difficulty stratification, and its effective application in curriculum learning validates the method's superiority. We open-source our pipeline to facilitate future research into scaling laws and automated curriculum learning.

However, we acknowledge certain limitations. Our scope is currently restricted to English math/code tasks, and the verification cost limits real-time use. Most critically, our pipeline faces the \textit{Verifier Paradox}: relying on a fixed-capacity verifier creates an epistemic ceiling, where valid problems exceeding the verifier's capabilities are at risk of being discarded as unsolvable. Future work must address this scalable oversight challenge.

\section*{Impact Statement}
Our work provides a foundational framework for scaling the difficulty of synthetic reasoning data while maintaining logical validity. By decoupling problem complexity from human curation, this research facilitates the development of more robust reasoning capabilities in AI systems across mathematical and programming domains. While this enables rapid progress in model performance, it also underscores the importance of integrating strict solvability constraints to prevent the degradation of data quality in automated training loops.

\clearpage

\bibliographystyle{plainnat}
\bibliography{main}

\clearpage

\beginappendix

\section{CoDiQ Pipeline: Case Study}
\label{appendix:case_study}

We demonstrate the CoDiQ pipeline through a complete workflow from an initial easy problem to iterative difficulty escalation, illustrating both successful upgrades and failure modes. Each generated problem undergoes solvability verification (Appendix~\ref{appendix:solvability_check}) and difficulty assessment (Appendix~\ref{appendix:llms_ranking}).

\subsection{Initial Problem}

\textbf{Problem Statement:} Count subsequences with an odd sum from array \texttt{nums}, returning the result modulo $10^9 + 7$.

\textbf{Example:} For \texttt{nums = [1,1,1]}, the answer is 4 (subsequences from positions: $\{0\}, \{1\}, \{2\}, \{0,1,2\}$, all with odd sums).

\textbf{Solution:} Simple DP tracking sum parity in $O(n)$ time.

\subsection{Round 1: Controlled Escalation}

\subsubsection{Upgraded Problem}

Count non-empty subsequences satisfying three simultaneous conditions:
\begin{enumerate}
    \item Sum is \textbf{odd}
    \item Length is \textbf{even}
    \item Sum $\bmod 3 = 1$
\end{enumerate}

\textbf{Difficulty Enhancement:} The upgrade introduces multi-dimensional state tracking, expanding the DP state space from 2 (sum parity) to $2 \times 2 \times 3 = 12$ states (sum parity, length parity, sum mod 3).

\subsubsection{Verification}

\begin{itemize}
    \item \textbf{Solvability Score:} 0.90
    \item \textbf{Time Complexity:} $O(12n) \approx 1.2 \times 10^6$ operations for $n = 10^5$ (feasible)
    \item \textbf{Solution Density:} $\sim 8\%$ of subsequences satisfy all conditions (non-trivial)
    \item \textbf{Solvability:} \textcolor{green}{PASS}
    \item \textbf{Difficulty:} \textcolor{green}{INCREASED}
\end{itemize}

\subsection{Round 2: Further Escalation}

\subsubsection{Upgraded Problem}

Count subsequences satisfying five conditions:
\begin{enumerate}
    \item Sum is \textbf{odd}
    \item Length is \textbf{even}
    \item Sum $\bmod 3 = 1$
    \item Sum $\bmod 5 = 2$
    \item Length $\bmod 4 = 2$
\end{enumerate}

\textbf{Mathematical Simplification:} By the Chinese Remainder Theorem (CRT), conditions 1, 3, and 4 can be unified:
\begin{align*}
\text{sum} \equiv 1 \pmod{2}, \quad \text{sum} \equiv 1 \pmod{3}, \quad \text{sum} \equiv 2 \pmod{5} \quad \Rightarrow \quad \text{sum} \equiv 7 \pmod{30}
\end{align*}

The effective state space becomes $30 \times 4 = 120$ states.

\subsubsection{Verification}

\begin{itemize}
    \item \textbf{Solvability Score:} 0.85
    \item \textbf{Time Complexity:} $O(120n) \approx 1.2 \times 10^7$ operations for $n = 10^5$ (acceptable)
    \item \textbf{Solution Density:} $\sim 0.83\%$ (still non-trivial)
    \item \textbf{Solvability:} \textcolor{green}{PASS}
    \item \textbf{Difficulty:} \textcolor{orange}{UNCHANGED}
\end{itemize}

\subsection{Round 3: Over-Escalation Failure}

\subsubsection{Upgraded Problem}

Count subsequences satisfying six conditions:
\begin{enumerate}
    \item Sum is \textbf{odd}
    \item Sum $\bmod 3 = 1$
    \item Sum $\bmod 5 = 2$
    \item Sum $\bmod 7 = 4$
    \item Sum $\bmod 11 = 6$
    \item Length $\bmod 8 = 2$ (which ensures even length)
\end{enumerate}

By CRT, conditions 1--5 unify to sum $\equiv c \pmod{2310}$ for some constant $c$, yielding a state space of $2310 \times 8 = 18{,}480$ states.

\subsubsection{Verification}

\begin{itemize}
    \item \textbf{Solvability Score:} 0.65
    \item \textbf{Solvability:} \textcolor{red}{FAIL}
    \item \textbf{Difficulty:} \textcolor{green}{INCREASED}
\end{itemize}

\paragraph{Failure Analysis:}

\textbf{1. Computational Infeasibility}
\begin{itemize}
    \item Time complexity: $O(18{,}480n) \approx 1.8 \times 10^9$ operations for $n = 10^5$
    \item Exceeds practical competitive programming limits (typically $\sim 10^8$--$10^9$ operations within time constraints)
\end{itemize}

\textbf{2. Solution Space Collapse} (Critical Issue)
\begin{itemize}
    \item While constraints are mathematically consistent via CRT, they create an extremely sparse solution space
    \item Probability that a random subsequence satisfies all conditions: $\approx \frac{1}{2310} \times \frac{1}{8} = \frac{1}{18{,}480}$
    \item Expected number of valid subsequences: $\frac{2^n}{18{,}480}$
    \item For $n \leq 14$: $\frac{2^{14}}{18{,}480} \approx 0.89 < 1$
    \item \textbf{Practical impact:} For typical inputs with small to moderate $n$, the answer is almost always 0, making the problem vacuously trivial
\end{itemize}

\paragraph{Pipeline Termination:}
The pipeline correctly terminates at Round 3, discarding $Q_3$ and outputting $\{Q_0, Q_1, Q_2\}$. Despite the increased theoretical difficulty, the problem becomes unsolvable due to computational infeasibility and solution space collapse, demonstrating the effectiveness of solvability verification in preventing quality degradation.
\section{CoDiQ Pipeline: Failure Type Analysis}
\label{appendix:failure_analysis}

To systematically understand the failure modes of the CoDiQ pipeline, we conduct a comprehensive clustering analysis on the collected failure reasons. Our analysis follows a three-stage hierarchical approach: initial K-means clustering, keyword extraction, and hierarchical merging with manual refinement.

\subsection{Clustering Methodology}

\textbf{Stage 1: K-means Pre-clustering.} We first apply K-means clustering to the failure reason descriptions to obtain an initial partitioning of the data. This pre-clustering step reduces computational complexity and provides a coarse-grained grouping of similar failure patterns.

\textbf{Stage 2: Keyword Extraction.} For each cluster obtained from K-means, we extract representative keywords using TF-IDF weighting. These keywords serve as semantic signatures that characterize the dominant failure patterns within each cluster, facilitating interpretability and subsequent hierarchical analysis.

\textbf{Stage 3: Hierarchical Clustering and Manual Refinement.} We then perform hierarchical clustering on the cluster centroids, leveraging the extracted keywords to compute semantic similarity. Finally, we manually merge related clusters and consolidate small clusters (containing fewer than a predefined threshold of samples) with their semantically nearest neighbors. This hybrid approach balances computational efficiency with semantic coherence.

\subsection{Failure Category Distribution}

Table~\ref{tab:failure_distribution} categorizes the identified failure modes. The analysis reveals two dominant distinct failure dynamics: \textbf{validity breaches} (Unsolvable) and \textbf{complexity degradation} (Difficulty Decreased).

\paragraph{Unsolvable Scenarios.} 
The majority of pipeline failures stem from fundamental deficits in problem formulation. Specifically, \textit{Definition \& Information Missing} combined with \textit{Constraints \& Logic Conflicts} collectively account for the lion's share of unsolvable cases. This indicates that the primary challenge lies not in parsing or formatting (which constitute a negligible fraction), but in the model's capacity to maintain semantic consistency and logical completeness during generation.

\paragraph{Difficulty Preservation.}
A critical observation is the prevalence of the \textit{Difficulty Decreased} category ($N=12,916$). In these instances, the generated problems remain solvable but fail to meet the intended cognitive demand. The high frequency of \textit{Constraint Simplification} and \textit{Numerical Range Reduction} suggests a model tendency towards "safe" or simplified generative paths, inadvertently pruning the solution space or removing key logical hurdles required for high-quality mathematical reasoning.

\begin{table}[htbp]
\centering
\caption{Distribution of Failure Cases in CoDiQ Pipeline}
\label{tab:failure_distribution}
\begin{tabular}{llr}
\toprule
\textbf{Failure Type} & \textbf{Failure Subtype} & \textbf{Count} \\
\midrule
Unsolvable & Definition \& Information Missing & 8,630 \\
 & Constraints \& Logic Conflicts & 8,142 \\
 & Computationally Infeasible & 1,948 \\
 & Implementation Details Missing & 1,926 \\
 & No Suitable Algorithm/Structure & 1,285 \\
 & Overly Complex & 1,104 \\
 & Requires Specific Capability & 726 \\
 & Parsing \& Rule Ambiguity & 611 \\
 & Number-Theoretic Constraints & 566 \\
 & Other & 1,184 \\
\midrule
Difficulty Decreased & Constraint Simplification & 3,245 \\
 & Numerical Range Reduction & 2,890 \\
 & Key Condition Removal & 2,654 \\
 & Solution Space Narrowing & 1,987 \\
 & Structural Simplification & 1,456 \\
 & Other & 684 \\
\addlinespace[0.5em]
\bottomrule
\end{tabular}
\end{table}

\section{ValueNetwork Training Detail}
\label{appendix:training}
\subsection{Dataset split} 
We compiled a labeled dataset by selecting samples from standard benchmarks~\cite{cobbe2021gsm8k, xia2025leetcodedataset} for the \textit{Easy} class and competition-level datasets~\cite{numina_math_datasets, hendrycksapps2021} for the \textit{Hard} class. We maintained an easy-to-hard ratio of 2:3 to prioritize the identification of challenging samples. We partition the compiled dataset into an 85:15 train-test split to ensure robust evaluation. 

\subsection{Training Data}
\subsubsection{Input Features}
For training data, we employ \textsc{Qwen3-8B} (in non-thinking mode) to capture generation dynamics. We define a sampling window from the last token of the question extending to $\min(4096, L_r)$ generated tokens. 

Within this window, we apply a \textit{quadratic sampling strategy} to select $K$ hidden states ($K=10$ for windows $>1024$, else $K=8$) at positions:

\begin{equation}
\label{eq:sampling}
p_i = \left\lfloor |W| \cdot \left(\frac{i}{K-1}\right)^2 \right\rfloor,\quad i = 0, 1, \dots, K-1.
\end{equation}

\subsubsection{Output Labels}
This strategy allocates higher sampling density to the onset of generation, capturing critical information for establishing the reasoning path. To mitigate stochasticity, scores are averaged over 5 independent passes.

% To train the Value Network, we formulate a binary classification task where the goal is to predict the correctness of a generation based on its internal representations. 

For each question, we generate a response using \textsc{Qwen3-8B} and assign a binary label $y \in \{0, 1\}$ based on the final answer's correctness. The input features $x$ are extracted via the quadratic sampling strategy (Eq.~\ref{eq:sampling}) applied to the first 4096 tokens.

% We employ a lightweight MLP to map these features to a scalar score, optimizing the binary cross-entropy loss. 

% At inference time, this predicted probability of correctness serves as a proxy for the LLM-perceived difficulty, where a lower score indicates a higher difficulty level. Detailed implementation and hyperparameters are provided in Appendix~\ref{appendix:training}.

\subsection{Network Architecture}
The Value Network is implemented as a lightweight Multi-Layer Perceptron (MLP) designed to project high-dimensional hidden states ($d_{in}=4096$) to a scalar correctness score. The architecture consists of an initial projection layer, Layer Normalization, GELU activation, and a final regression head.

This setup allows the network to minimize the discrepancy with the correctness label $y$ via a weighted binary cross-entropy objective, effectively estimating the likelihood of a successful generation solely from the reasoning dynamics captured in the early stages.

\subsection{Training Configuration}
The model is trained using the AdamW optimizer with a step learning rate scheduler. To address class imbalance, we apply a positive class weight in the loss function, dynamically calculated as the ratio of negative to positive samples. Complete hyperparameter settings are listed in Table~\ref{tab:hyperparams}.

\begin{table*}[t] % 如果是双栏论文请务必使用 table*，单栏论文使用 table 即可
\caption{Configuration and performance evaluation of the Value Network.}
\label{tab:valuenet_combined}
\vskip 0.15in
\begin{center}
\begin{small}
\begin{sc}

% --- 第一个表格 (Hyperparameters) ---
\begin{minipage}{0.48\textwidth}
    \centering
    \captionof{table}{Hyperparameter settings.}
    \label{tab:hyperparams}
    \vskip 0.1in % 微调 caption 和表格的间距
    \begin{tabular}{ll}
    \toprule
    \textbf{Hyperparameter} & \textbf{Value} \\
    \midrule
    Input Dim ($d_{in}$) & 4096 \\
    Hidden Dim & 512 \\
    Batch Size & 512 \\
    Learning Rate & $1 \times 10^{-4}$ \\
    Weight Decay & $1 \times 10^{-2}$ \\
    Dropout & 0.3 \\
    Optimizer & AdamW \\
    Scheduler & StepLR ($\gamma=0.8$) \\
    Max Epochs & 30 \\
    Split & 85\% / 15\% \\
    \bottomrule
    \end{tabular}
\end{minipage}
\hfill % 填充中间空白，把两个表格推向两边
% --- 第二个表格 (Performance) ---
\begin{minipage}{0.48\textwidth}
    \centering
    \captionof{table}{Performance on held-out test set.}
    \label{tab:valuenet_perf}
    \vskip 0.1in
    \begin{tabular}{lc}
    \toprule
    \textbf{Metric} & \textbf{Value} \\
    \midrule
    Accuracy & 72.52\% \\
    Precision & 54.21\% \\
    Recall & 95.62\% \\
    F1 Score & 69.20\% \\
    \textbf{ROC-AUC} & \textbf{84.84\%} \\
    PR-AUC & 65.77\% \\
    % & \\ % 稍微占位以对齐行数，或者可以删掉这行让表格自然高度
    % & \\
    % & \\
    % & \\
    \bottomrule
    \end{tabular}
\end{minipage}

\end{sc}
\end{small}
\end{center}
\vskip -0.1in
\end{table*}

\subsection{Performance Evaluation}
\label{appendix:valuenetwork_performance}
We evaluate the trained Value Network on the held-out test set (15\% split). As shown in Table~\ref{tab:valuenet_perf}, the model achieves an \textbf{ROC-AUC of 84.84\%}, demonstrating robust discriminative power in distinguishing correct reasoning paths from incorrect ones despite the challenging nature of the dataset.

It is worth noting that our training strategy prioritizes identifying all potential correct answers. This is reflected in the \textbf{high Recall of 95.62\%}, which ensures that the Value Network successfully preserves valid reasoning paths. While this focus on coverage results in a moderate Precision (54.21\%) due to the trade-off inherent in class-weighted training, the high ROC-AUC indicates that the predicted scores effectively rank correct generations higher, making the model reliable for difficulty estimation and filtering.

\section{CoDiQ-Bench Selection Criteria}
\label{appendix:bench_selection}
To ensure the quality and reliability of our benchmark, we establish three primary criteria for data selection:

\textbf{Solvability:} We verify that each problem is well-defined and admits at least one valid solution, ensuring the benchmark's validity and fairness.

\textbf{Difficulty Level:} We assess whether the difficulty level is appropriate for the intended evaluation purpose, maintaining a balanced distribution across different complexity levels.

\textbf{Quality Assessment:} We conduct rigorous quality checks to ensure that all selected problems meet acceptable standards in terms of clarity, correctness, and relevance.

\section{Statistics of CoDiQ-Corpus}
\label{appendix:codiq_corpus}

\begin{table*}[htbp]
\centering
\caption{Dataset statistics of CoDiQ-Corpus.}
\begin{tabular}{lcccccc}
\toprule
\textbf{Dataset} & \multicolumn{3}{c}{\textbf{Question Tokens Length}} & \textbf{AVG Round} & \textbf{Category} & \textbf{Sequences} \\
\cmidrule(lr){2-4}
& \textbf{Minimum} & \textbf{Maximum} & \textbf{Average} & & & \\
\midrule
Math12K~\cite{hiyouga_math12k} & 38 & 7,829 & 995.4 & 4.7 & Math & 11,764 \\
GSM8K~\cite{cobbe2021gsm8k} & 52 & 6,896 & 1,093.7 & 4.5 & Math & 8,685 \\
SVAMP~\cite{ChilleD_SVAMP} & 172 & 3,992 & 971.3 & 3.3 & Math & 804 \\
ASDiv~\cite{yimingzhang_asdiv} & 55 & 4,703 & 1,013.1 & 4.7 & Math & 1,480 \\
CodeAlpaca20K~\cite{codealpaca} & 70 & 7,174 & 1,106.1 & 3.8 & Code & 17,845 \\
LeetCodeDataset~\cite{xia2025leetcodedataset} & 254 & 4,365 & 1,281.0 & 3.8 & Code & 2,027 \\
MBPP~\cite{austin2021program} & 52 & 3,440 & 1,000.4 & 3.4 & Code & 876 \\
DS-1000~\cite{Lai2022DS1000} & 192 & 4,138 & 1,240.7 & 3.2 & Code & 972 \\
\midrule
Total & 38 & 7,829 & 1,073.0 & 4.2 & - & 44,453 \\
\bottomrule
\end{tabular}
\label{tab:dataset_details}
\end{table*}

We employ CoDiQ-Gen-8B following the CoDiQ Pipeline (Section~\ref{sec:codiq_pipeline}) to transform eight diverse mathematical and programming datasets into the more challenging CoDiQ-Corpus, which comprises approximately 44,453 question sequences with progressive difficulty from easy to hard. The detailed distribution is presented in Table~\ref{tab:dataset_details}.
\section{Scaling Tendency Analysis Details}
\label{appendix:scaling_tendency}
\begin{figure}[!tb]
    \centering
    \setlength{\tabcolsep}{2pt}  % 减小列间距
    \renewcommand{\arraystretch}{0.2}  % 减小行间距
    \begin{tabular}{cc}
        \includegraphics[width=0.48\linewidth]{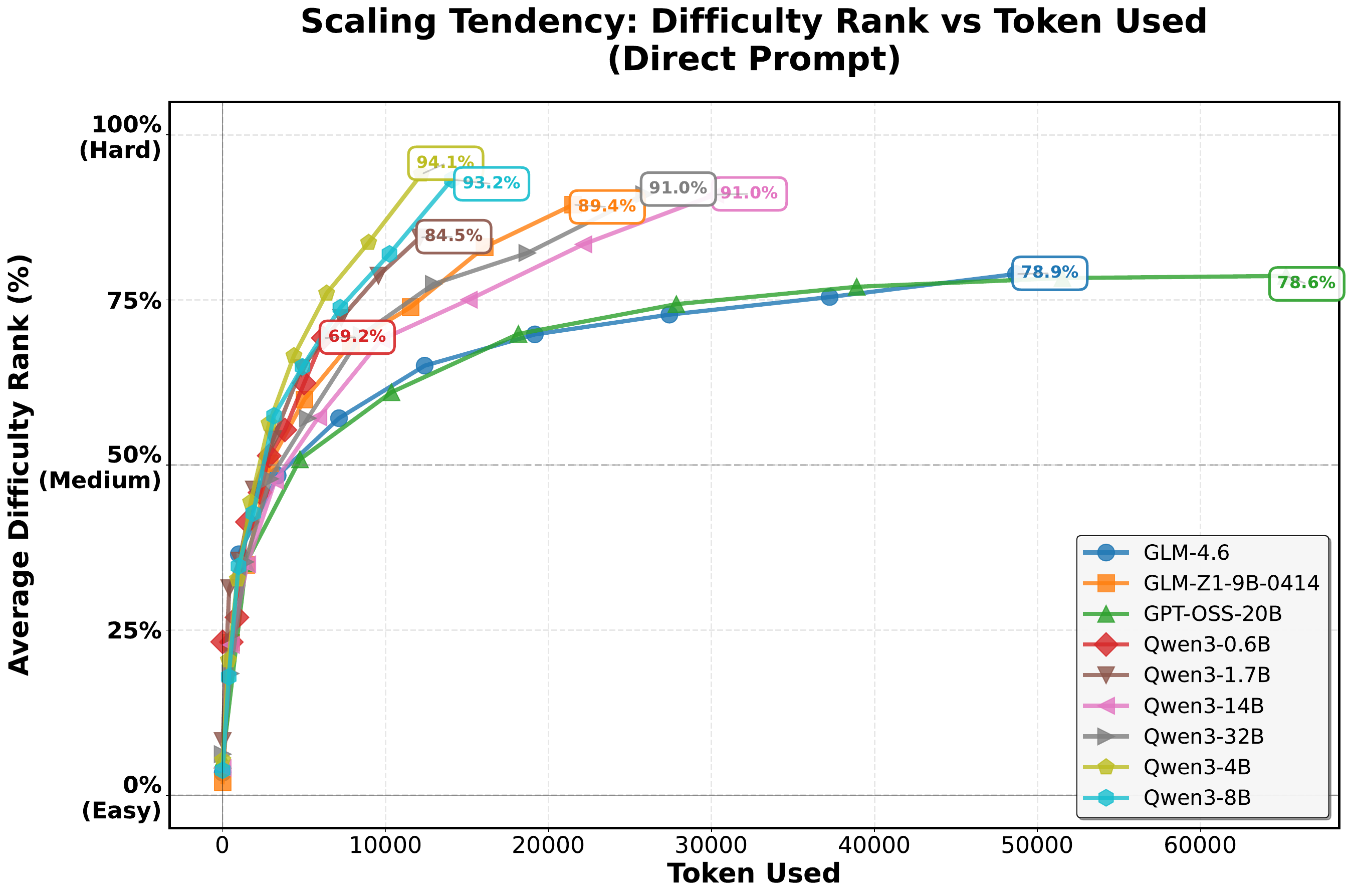} &
        \includegraphics[width=0.48\linewidth]{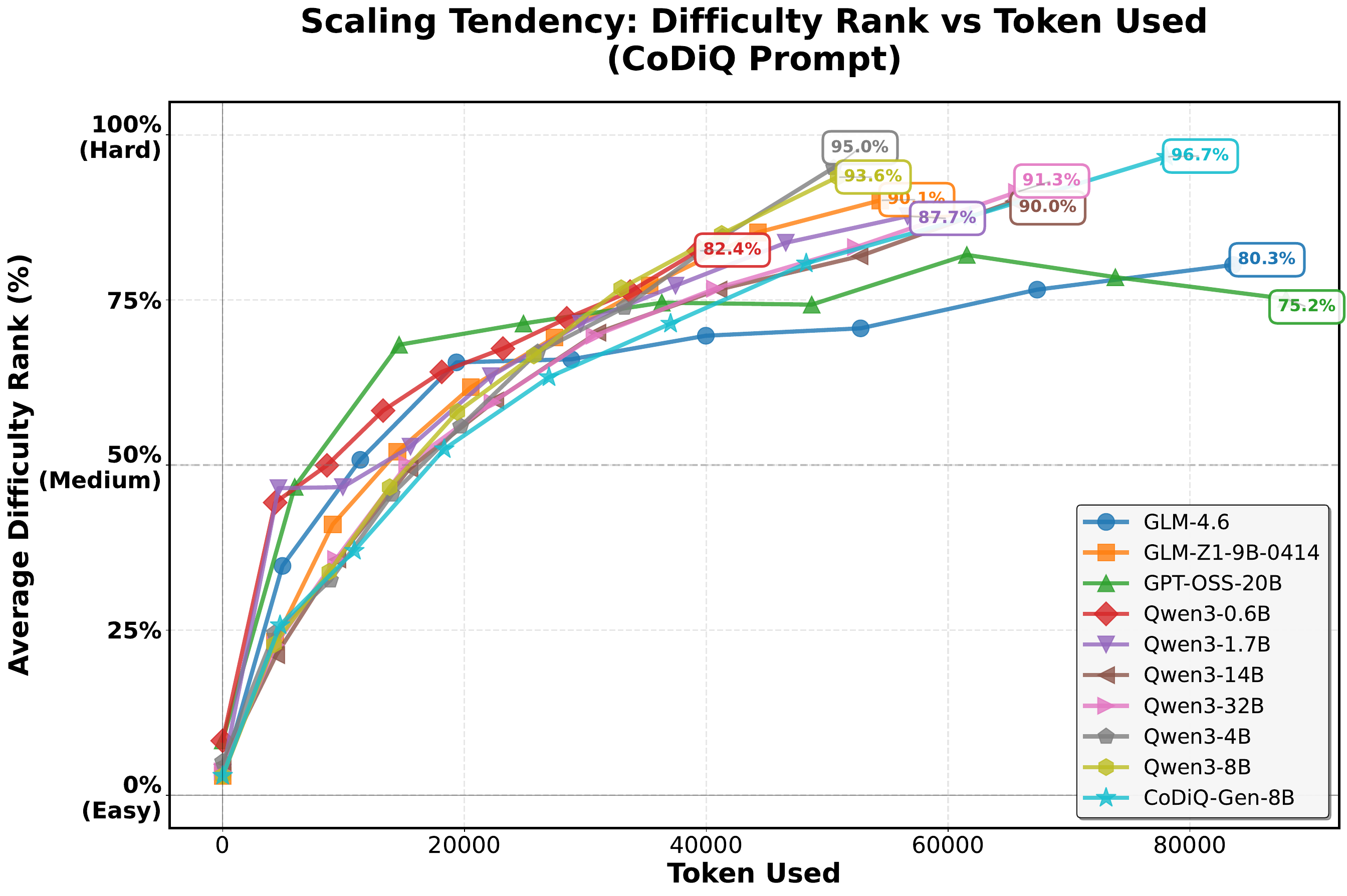} \\
        \\[1em]
        \includegraphics[width=0.48\linewidth]{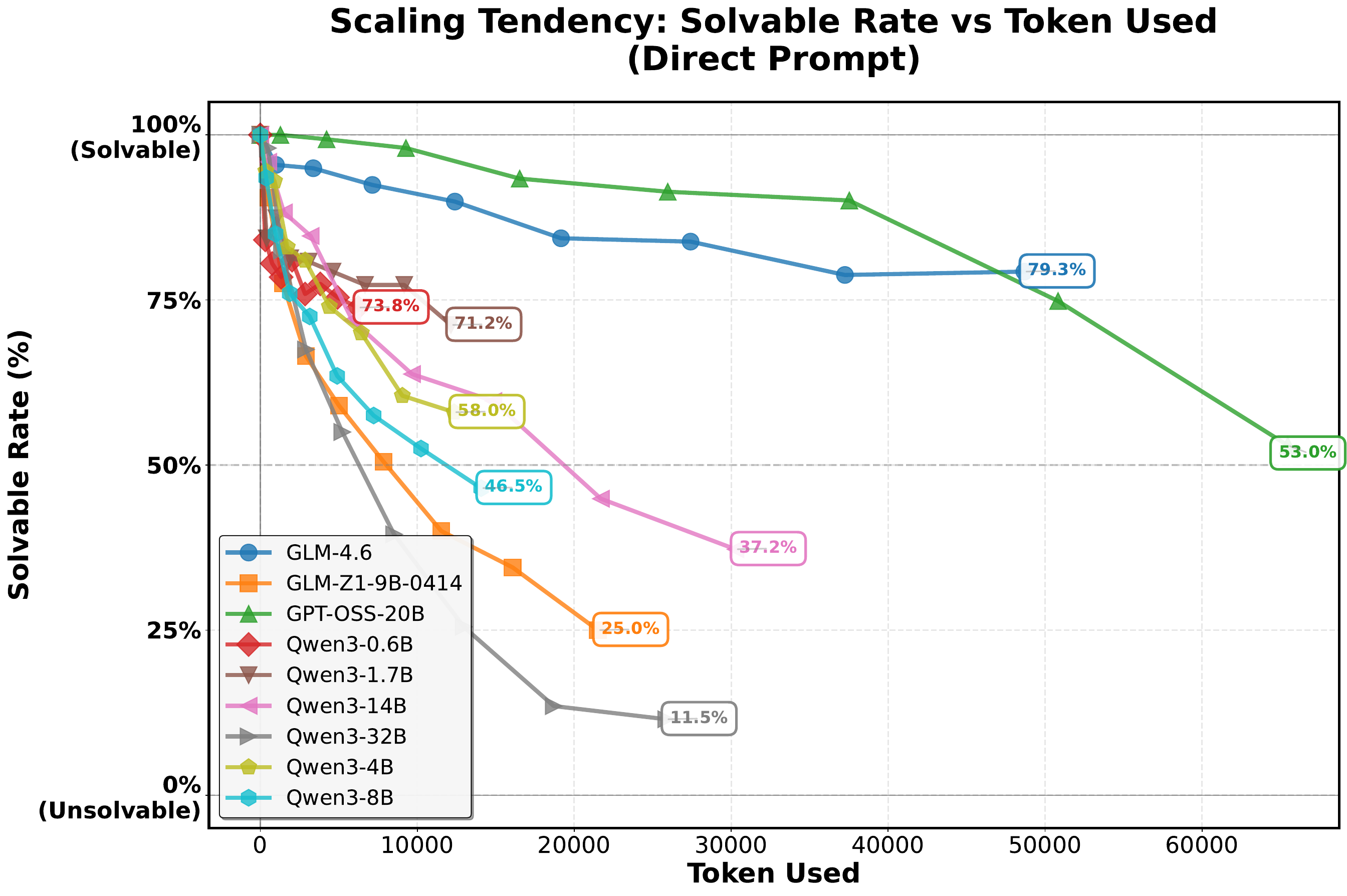} &
        \includegraphics[width=0.48\linewidth]{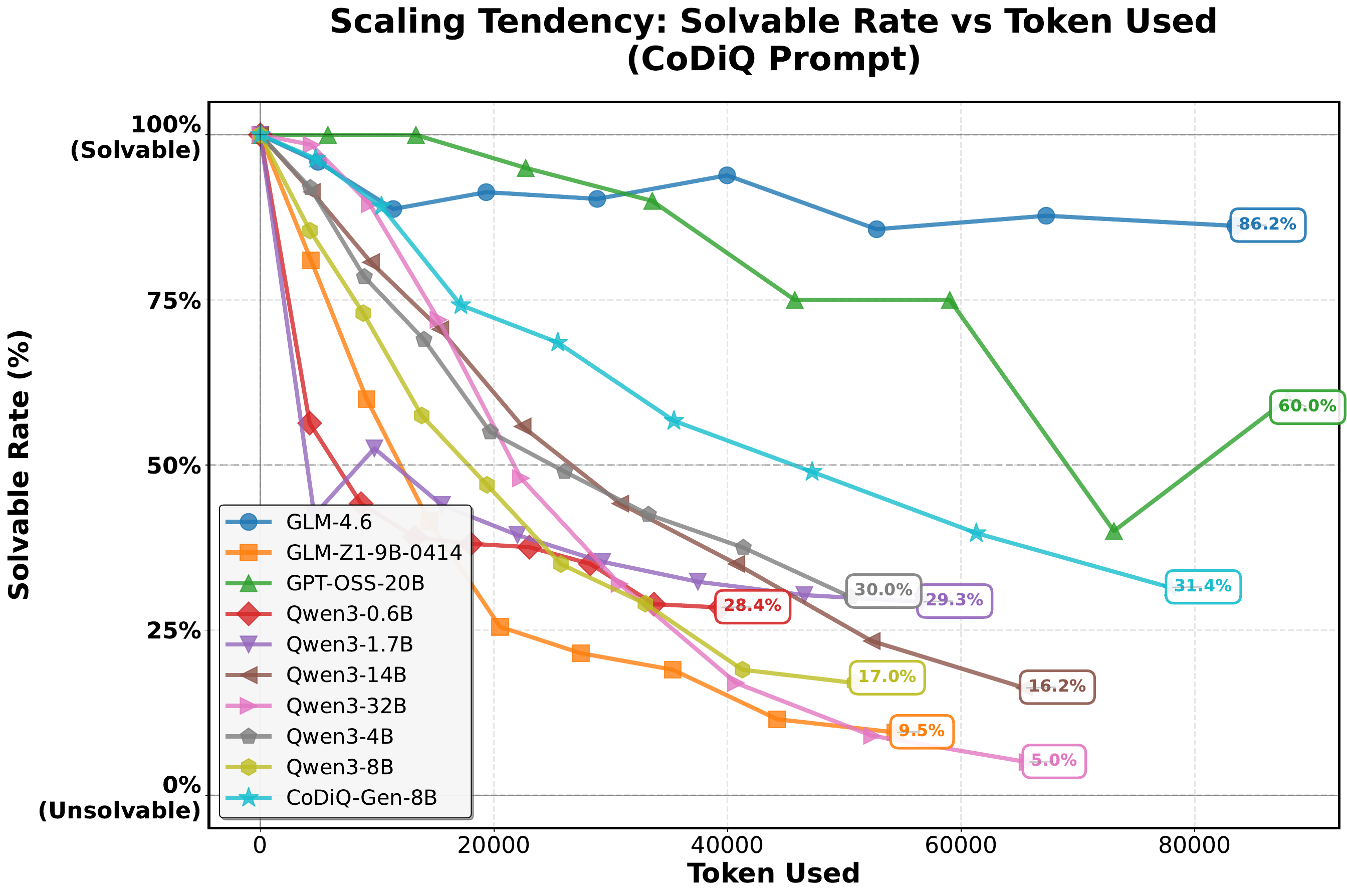} \\
    \end{tabular}
    \caption{\textbf{Complete Scaling Analysis on CoDiQ-Bench.} Normalized average difficulty ranking (top row) and solvable rate (bottom row) of questions generated by all evaluated Long-CoT models across 8 rounds, using Direct Prompt (left) and CoDiQ Prompt (right). Higher rankings indicate higher question difficulty; higher rates indicate better question quality.}
    \label{fig:scaling_tendency_analysis}
\end{figure}

This section presents the complete scaling tendency analysis with all evaluated models. Figure~\ref{fig:scaling_tendency_analysis} shows the full results of difficulty and solvability scaling across 8 generation rounds for all Long-CoT models under both Direct Prompt and CoDiQ Prompt settings.

The complete results reveal consistent scaling patterns across all models: (1) increased reasoning computation correlates with higher problem difficulty, and (2) a trade-off exists between difficulty and solvability, with larger models maintaining better balance between the two metrics.
\section{CoDiQ-Corpus Quality Criteria}
\label{appendix:data_quality_criteria}

We establish rigorous criteria to assess the quality and solvability of problems in CoDiQ-Corpus. Three PhD-level domain experts independently evaluate 300 randomly sampled problems following these standardized guidelines:

\subsection{Information Completeness}
\begin{itemize}
    \item \textbf{Sufficient Parameters}: All necessary numerical values, variables, and constraints are explicitly provided.
    \item \textbf{Clear Objectives}: The problem goal is unambiguous and well-defined.
    \item \textbf{Complete Context}: No truncation or missing problem statements.
\end{itemize}

\subsection{Logical Consistency}
\begin{itemize}
    \item \textbf{Non-contradictory Conditions}: All given constraints are mutually consistent.
    \item \textbf{Valid Premises}: For logical problems, premises are sufficient to support conclusions.
    \item \textbf{Feasible Solutions}: The problem admits at least one valid solution path.
\end{itemize}

\subsection{Problem Well-definedness}
\begin{itemize}
    \item \textbf{Determinable Answer}: The answer can be uniquely determined or bounded within a reasonable range.
    \item \textbf{Appropriate Scope}: The problem complexity matches its stated domain and difficulty level.
    \item \textbf{Standard Formulation}: Follows conventional mathematical or logical notation.
\end{itemize}

\subsection{Evaluation Protocol}
Each expert assigns a binary solvability label (solvable/unsolvable) with confidence scores. A problem is marked as \textbf{solvable} only when at least two experts agree. Disagreements are resolved through discussion. The inter-annotator agreement (Fleiss' $\kappa$) reaches 0.78, indicating substantial consensus.
\section{Curriculum learning Detail}
\label{appendix:curriculum_learning_detail}

\subsection{Training Data Selection for Curriculum Learning}
\label{appendix:data_selection}

To validate the effectiveness of CoDiQ-Corpus for curriculum learning, we carefully select question sequences with progressive difficulty structures. Specifically, we sample 480 question sequences from CoDiQ-Corpus where each sequence length $|S| \geq 3$, forming the curriculum learning dataset:

\vspace{-2mm}
\begin{small}
\begin{equation}
\mathcal{D}_{\text{curriculum}} = \{S_n\}_{n=1}^{480}, \quad |S_n| \geq 3
\label{eq:curriculum_dataset}
\end{equation}
\end{small}
\vspace{-2mm}

For each sequence $S_n = \{q_0^n, q_1^n, q_2^n, \ldots, q_{|S_n|-1}^n\}$ with progressive difficulty, we construct three training stages with increasing complexity:

\begin{itemize}
    \item \textbf{Level 1 (L1):} Contains all initial questions $q_0^n$ from each sequence, representing the starting point of each difficulty progression.
    \item \textbf{Level 2 (L2):} Randomly samples one question from intermediate positions $\{q_1^n, q_2^n, \ldots, q_{|S_n|-2}^n\}$ for each sequence, capturing mid-stage complexity.
    \item \textbf{Level 3 (L3):} Contains all final questions $q_{|S_n|-1}^n$ from each sequence, representing the highest difficulty level within each progression.
\end{itemize}

Formally, the data selection strategy is defined as:

\vspace{-2mm}
\begin{small}
\begin{align}
\mathcal{L}_1 &= \{q_0^n \mid S_n \in \mathcal{D}_{\text{curriculum}}\} \label{eq:level1} \\
\mathcal{L}_2 &= \{\text{random}(\{q_i^n\}_{i=1}^{|S_n|-2}) \mid S_n \in \mathcal{D}_{\text{curriculum}}\} \label{eq:level2} \\
\mathcal{L}_3 &= \{q_{|S_n|-1}^n \mid S_n \in \mathcal{D}_{\text{curriculum}}\} \label{eq:level3}
\end{align}
\end{small}
\vspace{-2mm}

This design ensures a clear difficulty progression: $\text{Difficulty}(\mathcal{L}_1) < \text{Difficulty}(\mathcal{L}_2) < \text{Difficulty}(\mathcal{L}_3)$. The sample distribution across levels follows the ratio $|\mathcal{L}_1| : |\mathcal{L}_2| : |\mathcal{L}_3| = 2:2:1$, achieved by duplicating $\mathcal{L}_1$ and $\mathcal{L}_2$ during training to balance exposure to different difficulty levels. This ratio is designed to provide sufficient foundational learning before progressing to more challenging problems, following curriculum learning principles~\cite{bengio2009curriculum}.

For the baseline model Qwen3-RL-4B, we use the original untransformed datasets (before applying the CoDiQ Pipeline) as training data, maintaining the same total number of training samples to ensure fair comparison. This allows us to isolate the impact of progressive difficulty transformation on model performance.

\noindent\textbf{Training Schedule:} Models are trained sequentially through three stages:
\begin{enumerate}
    \item CoDiQ-L$_1$-4B: Trained on $\mathcal{L}_1$ (starting level)
    \item CoDiQ-L$_2$-4B: Initialized from CoDiQ-L$_1$-4B, further trained on $\mathcal{L}_2$ (intermediate level)
    \item CoDiQ-L$_3$-4B: Initialized from CoDiQ-L$_2$-4B, further trained on $\mathcal{L}_3$ (advanced level)
\end{enumerate}

\subsection{Reward Signal Design}
\label{appendix:reward_model}

We design a multi-dimensional reward function to evaluate answer quality by prompting Qwen3-32B as an expert evaluator. The reward signal $r \in [0, 1]$ is computed based on four key dimensions:

\noindent\textbf{Evaluation Dimensions:}
\begin{itemize}
    \item \textbf{Problem Resolution} ($s_{\text{pr}}$): Measures how completely the answer addresses all aspects of the question (0.0-1.0).
    \item \textbf{Reasoning Correctness} ($s_{\text{rc}}$): Evaluates the correctness and coherence of the reasoning process (0.0-1.0).
    \item \textbf{Information Completeness} ($s_{\text{ic}}$): Assesses whether all necessary information, steps, and explanations are included (0.0-1.0).
    \item \textbf{Accuracy} ($s_{\text{acc}}$): Measures factual correctness, calculation accuracy, and conceptual clarity (0.0-1.0).
\end{itemize}

The reward function aggregates these dimensions with carefully tuned weights optimized for high-difficulty mathematical reasoning tasks:

\vspace{-2mm}
\begin{small}
\begin{equation}
r = w_{\text{pr}} \cdot s_{\text{pr}} + w_{\text{rc}} \cdot s_{\text{rc}} + w_{\text{ic}} \cdot s_{\text{ic}} + w_{\text{acc}} \cdot s_{\text{acc}}
\label{eq:reward_aggregation}
\end{equation}
\end{small}
\vspace{-2mm}

where the default weights are set as:
\vspace{-2mm}
\begin{small}
\begin{equation}
w_{\text{pr}} = 0.20, \quad w_{\text{rc}} = 0.35, \quad w_{\text{ic}} = 0.25, \quad w_{\text{acc}} = 0.20
\label{eq:reward_weights}
\end{equation}
\end{small}
\vspace{-2mm}

This configuration emphasizes reasoning quality (35\%) and information completeness (25\%), which are critical for complex problem-solving. The evaluation prompt instructs Qwen3-32B to assess each dimension independently using continuous scores and return results in JSON format. Special handling is applied for edge cases, such as correctly identifying unsolvable problems, which receives high problem resolution scores (0.8-1.0) despite not providing a numerical solution.

To ensure evaluation quality, we implement automatic validation of the returned scores, retry mechanisms (up to 3 attempts), and text truncation to handle long inputs (max 4096 tokens for questions, 16384 tokens for answers). The confidence score returned by the evaluator helps identify uncertain assessments for potential manual review.
\section{Instruction for LLMs Ranking}
\label{appendix:llms_ranking}
\begin{promptbox}[Instruction for LLMs Ranking]
You are an expert in assessing question difficulty. Evaluate questions based on:

\begin{enumerate}
    \item Knowledge Complexity: Number and depth of concepts required
    \item Cognitive Load: Reasoning levels and abstract thinking needed
    \item Computational Complexity: Steps and calculations involved
    \item Traps and Common Mistakes: Hidden pitfalls in the question
    \item Integration Skills: Cross-domain knowledge application required
\end{enumerate}

Your task is to group questions by difficulty level and sort groups from easiest to hardest.\\

\textbf{Important:} Questions with the SAME difficulty level should be grouped together.\\

Analyze each question carefully and return them grouped by difficulty level.\\

\textbf{Output format requirements:}
\begin{itemize}
    \item Return ONLY a valid JSON object with TWO fields:
    \item \textbf{"result"}: A list of lists (groups), each containing question indices of the SAME difficulty level
    \item \textbf{"reason"}: A list of strings, where \texttt{reason[i]} explains why questions in \texttt{result[i]} share the same difficulty
    \item Groups in both arrays should be ordered from easiest to hardest
    \item The length of "result" and "reason" arrays MUST be identical
    \item Use 0-based indexing matching the input order
\end{itemize}

\textbf{Example output format:}
\begin{lstlisting}
{
  "result": [[1, 3], [0], [2, 4]],
  "reason": [
    "Both require only basic arithmetic operations with single-step reasoning",
    "Multi-step algebraic manipulation with intermediate concepts",
    "Complex integration of advanced concepts and non-obvious strategies"
  ]
}
\end{lstlisting}

This means: 
\begin{itemize}
    \item Questions 1 and 3 are easiest (Group 0) - they both involve only basic arithmetic and single-step reasoning
    \item Question 0 is medium difficulty (Group 1) - it requires multi-step algebraic manipulation with intermediate concepts
    \item Questions 2 and 4 are hardest (Group 2) - they both demand complex integration of advanced concepts and non-obvious strategies
\end{itemize}

\textbf{Important:} 
\begin{itemize}
    \item Each reasoning string should explain the COMMON difficulty characteristics that unite all questions in the corresponding group
    \item Ensure \texttt{reason[i]} corresponds to \texttt{result[i]} for all groups
\end{itemize}

Please group the following questions by difficulty level and sort groups from easiest to hardest:

\{\text{questions}\}\\

Return the result as a JSON object with format:
\begin{lstlisting}
{
  "result": [[indices of easiest group], [indices of next group], ...],
  "reason": ["reasoning for group 0", "reasoning for group 1", ...]
}
\end{lstlisting}
\end{promptbox}

% \textbf{Output format requirements:}
% \begin{itemize}
%     \item Return ONLY a valid JSON object
%     \item The JSON must have a "result" field containing a list of lists (groups)
%     \item Each inner list contains question indices of the SAME difficulty level
%     \item Groups should be ordered from easiest to hardest
%     \item Use 0-based indexing matching the input order
% \end{itemize}

% \textbf{Example output format:}
% \begin{lstlisting}
% {"result": [[1, 3], [0], [2, 4]]}
% \end{lstlisting}

% This means: 
% \begin{itemize}
%     \item Questions 1 and 3 are easiest (same difficulty)
%     \item Question 0 is medium difficulty
%     \item Questions 2 and 4 are hardest (same difficulty)
% \end{itemize}

% Please group the following questions by difficulty level and sort groups from easiest to hardest:

% \{\text{questions}\}\\

% Return the result as JSON with format: 
% \begin{lstlisting}
% {{"result": [[indices of easiest group], [indices of next group], ...]}}
% \end{lstlisting}

% \end{promptbox}
\section{Instructions for Solvability Check}
\label{appendix:solvability_check}
\begin{promptbox}[Instructions for Solvability Check]
You are an expert in analyzing mathematical and logical problems. Your task is to determine whether a given question is solvable.

A question is considered \textbf{SOLVABLE} if:
\begin{enumerate}
    \item It provides all necessary information and conditions
    \item The problem is well-defined with clear objectives
    \item It has a determinable answer (even if complex)
    \item The constraints are consistent (not contradictory)
\end{enumerate}

A question is considered \textbf{UNSOLVABLE} if:
\begin{enumerate}
    \item Missing critical information or parameters
    \item Contains contradictory conditions
    \item The problem statement is ambiguous or unclear
    \item Asks for information that cannot be determined from given data
    \item The question is incomplete or truncated
\end{enumerate}

\textbf{Important Guidelines:}
\begin{itemize}
    \item Be strict but reasonable in your judgment
    \item Consider if a reasonable person could solve the problem with the given information
    \item For mathematical problems, check if all necessary values are provided
    \item For logical problems, verify if the premises are sufficient for the conclusion
\end{itemize}

\textbf{Output format requirements:}
\begin{itemize}
    \item Return ONLY a valid JSON object
    \item Must have exactly these fields:
    \begin{itemize}
        \item "solvable": boolean (true/false)
        \item "confidence": number (0.0-1.0, your confidence in the judgment)
        \item "reason": string (brief explanation in English, max 200 characters)
        \item "missing\_info": list of strings (what information is missing, empty list if solvable)
    \end{itemize}
\end{itemize}

\textbf{Example Outputs:}
\begin{lstlisting}
{"solvable": true, "confidence": 0.95, "reason": "All necessary parameters provided, problem is well-defined", "missing_info": []}
{"solvable": false, "confidence": 0.85, "reason": "Missing the radius value needed to calculate circle area", "missing_info": ["radius"]}
\end{lstlisting}

\end{promptbox}
\section{Instruction for Direct Prompt}
\label{appendix:direct_prompt}
\begin{promptbox}[Instruction for Direct Prompt]

\textbf{\# Problem Difficulty Upgrade Generator}

\textbf{\#\# Task Description}

You are an expert competitive programming problem creator. Your task is to take a given problem and create a significantly more challenging, competition-level version.\\

\textbf{\#\# Input}

\textbf{Original Problem:}\\
\{\text{original\_problem}\}\\

\textbf{\#\# Output Format}\\
Return ONLY the new upgraded problem, nothing else.\\

[Your upgraded competitive programming problem here]

\end{promptbox}
\section{Instruction for CoDiQ Prompt}
\label{appendix:codiq_prompt}
\begin{promptbox}[Instruction for CoDiQ Prompt]

\textbf{\# Problem Difficulty Upgrade Generator}

\textbf{\#\# Task Description}

You are an expert competitive programming problem creator. Your task is to take a given problem and create a significantly more challenging, competition-level version by strategically adding difficulty elements that test deeper understanding and more complex reasoning.\\

\textbf{\#\# Design Standards (Mandatory Quality Check)}\\
To ensure the upgraded problem is competition-worthy, you must strictly adhere to these principles:
\begin{enumerate}
    \item \textbf{Deep Synthesis}: The difficulty element must naturally intertwine with the original logic. The solution should feel like a single cohesive challenge, not a "patchwork".
    \item \textbf{Multi-Step Reasoning}: The solution must require 2-3 non-trivial intermediate logical jumps. The solver must derive lemmas or intermediate states before applying standard algorithms.
    \item \textbf{No Trivial Upgrades}: Avoid simply increasing N to $10^5$ if the logic remains $O(N)$. The upgrade must force a change in complexity class (e.g., from Greedy to Flow, from Simulation to Matrix Exponentiation).
    \item \textbf{Disguise \& Abstraction}: (If applicable) Hide the core theorem or data structure behind a unique story or abstract mathematical setting. Never explicitly name the required algorithm.
\end{enumerate}

\textbf{\#\# Difficulty Elements Library (Select 1-2 distinct elements)}\\
% 1
\textbf{\#\#\# Category A: }\\
\textbf{Dimensionality \& Constraints}\\
\textbf{Best for:} Array/Sequence/Tree problems with simple naive solutions (e.g., $O(N)$, $O(N^2)$, or $O(N^3)$).\\
\textbf{Avoid when:} Original problem already requires logarithmic or sublinear complexity.\\
\textbf{Description:} Explode the data scale or dimensionality to invalidate simple simulation or brute force.\\
\textbf{Core Strategy:}
\begin{enumerate}
    \item \textbf{Identify} the naive complexity (e.g., $O(N)$, $O(N^2)$, or $O(N^3)$).
    \item \textbf{Impose} constraints that compel a superior complexity class (e.g., $O(\log N)$ or $O(N \log N)$).
    \item \textbf{Introduce} dynamic updates, higher-dimensional spaces, or multiple query types to break linear scans.
\end{enumerate}

\textbf{Examples:}
\begin{lstlisting}
[
    {
        "original": "Given an array of size N (N≤1000), find the sum of elements in range [L, R].",
        "upgrade": "Given an array of size N (N≤10^5), handle M (M≤10^5) operations: 1. Update range [L, R] by adding V. 2. Query sum of range [L, R]. (Requires Segment Tree with Lazy Propagation)"
    },
    {
        "original": "Given a grid, find the shortest path from (0,0) to (R,C) avoiding obstacles.",
        "upgrade": "Given a grid where obstacles appear and disappear at specific time intervals modulo K. Find the shortest path. (Requires BFS in State Space (x, y, time%K))"
    },
    {
        "original": "Find the maximum value in an array.",
        "upgrade": "Given a tree with N nodes (N≤10^5), support path updates (add value V to all nodes on path u-v) and path maximum queries. (Requires Heavy-Light Decomposition)"
    },
    {
        "original": "Given a set of points, find the two closest points.",
        "upgrade": "Given a set of points in 3D space, find the size of the largest subset where every pair has Manhattan distance > D. (Requires Coordinate Transformation + Data Structures)"
    },
    {
        "original": "Check if a string S contains pattern P.",
        "upgrade": "Given a text S and K patterns. Support dynamic insertion of new patterns and query if any pattern appears in S. (Requires Aho-Corasick Automaton or Suffix Structures)"
    }
]
\end{lstlisting}

% 2
\textbf{\#\#\# Category B: }\\
\textbf{Mathematical Abstraction}\\
\textbf{Best for:} Problems that can be reframed into mathematical structures, e.g., simulation or iterative problems with clear patterns.\\
\textbf{Avoid when:} The problem is already focused on advanced specialized theorems or complex data structures.\\
\textbf{Description:} Transform a procedural or descriptive problem into a formal model, e.g., using number theory, combinatorics, or game theory.\\
\textbf{Core Strategy:}
\begin{enumerate}
    \item \textbf{Increase} constraints to push beyond computational limits (e.g., $N \ge 10^{18}$), making simple iteration or simulation impossible.
    \item \textbf{Force} the discovery of underlying structures, such as closed-form formulas, recurrence relations, or invariant properties.
    \item \textbf{Introduce} formal constraints (e.g., modular arithmetic, coordinate systems) that require rigorous mathematical modeling.
\end{enumerate}

\textbf{Anti-pattern:} Simply making N large without ensuring a mathematical insight exists is not valid.\\
\textbf{Examples:}
\begin{lstlisting}
[
    {
        "original": "Simulate a process where bacteria double every hour. Find count at hour N (N≤50).",
        "upgrade": "Bacteria have a complex growth rule F(n) = a*F(n-1) + b*F(n-2). Find count at hour N (N≤10^18) modulo 10^9+7. (Requires Matrix Exponentiation)"
    },
    {
        "original": "Given N items, in how many ways can you pick K items?",
        "upgrade": "Given N items with specific color constraints, calculate the number of ways to pick K items modulo 10^9+7 where N is up to 10^9. (Requires Lucas Theorem or Generating Functions)"
    },
    {
        "original": "Two players take turns removing 1-3 stones. Who wins?",
        "upgrade": "Played on a graph with N≤10^5 nodes. A token moves along edges. A player loses if they cannot move. The graph has cycles. (Requires Game Theory on Graphs / Sprague-Grundy with loop handling)"
    },
    {
        "original": "Calculate the Greatest Common Divisor (GCD) of two numbers.",
        "upgrade": "Calculate the sum of GCD(i, j) for all 1 ≤ i, j ≤ N where N≤10^7. (Requires Euler Totient Function / Mobius Inversion)"
    },
    {
        "original": "Find the area of a polygon given integer coordinates.",
        "upgrade": "Given N lines in the plane, find the area of their union region accurately. Handle parallel and concurrent lines. (Requires Integration logic or Green's Theorem application)"
    }
]
\end{lstlisting}

% 3
\textbf{\#\#\# Category C: }\\
\textbf{Inverse \& Constructive}\\
\textbf{Best for:} Problems with well-defined algorithms and a clear "input → algorithm → output" flow.\\
\textbf{Avoid when:} The original problem's core challenge is already in the "design/construction" phase rather than "computation/processing" (i.e., problems that lack a standard algorithm to reverse).\\
\textbf{Description:} Instead of asking for the result of a process, ask for the input that produces a specific result.\\
\textbf{Core Strategy:}
\begin{enumerate}
    \item \textbf{Reverse}the problem direction: from "Given X, find Y" to "Construct X such that Y holds".
    \item \textbf{Require} understanding of structural properties (e.g., what makes a graph have a specific flow?).
    \item \textbf{Add} multiple constraints to make construction non-trivial.
\end{enumerate}

\textbf{Examples:}
\begin{lstlisting}
[
    {
        "original": "Given a graph, find the shortest path from A to B.",
        "upgrade": "Construct a graph with N vertices and M edges such that the shortest path from 1 to N is exactly L, and the MST weight is exactly W."
    },
    {
        "original": "Sort an array using QuickSort.",
        "upgrade": "Construct a permutation of size N that causes a standard QuickSort implementation (with first element as pivot) to hit its worst-case O(N^2) time complexity."
    },
    {
        "original": "Given a binary tree, print its pre-order traversal.",
        "upgrade": "Given the pre-order and post-order traversals, reconstruct all possible binary trees. Determine if the solution is unique or count how many such trees exist."
    },
    {
        "original": "Check if a string is a palindrome.",
        "upgrade": "Construct a string of length N containing exactly K distinct palindromic substrings. Prove that no such string exists if K exceeds a certain bound."
    },
    {
        "original": "Find the maximum flow in a network.",
        "upgrade": "Given a desired max flow value F, construct a network with minimum edges that achieves this flow, subject to capacity constraints on each edge."
    }
]
\end{lstlisting}

% 4
\textbf{\#\#\# Category D: }\\
\textbf{State Explosion}\\
\textbf{Best for:} Problems with simple, polynomial DP states that can be enriched.\\
\textbf{Avoid when:} The original state space is already exponential (e.g., TSP); adding dimensions would make it computationally infeasible.\\
\textbf{Description:} Add complex dependencies or history requirements that necessitate advanced Dynamic Programming or Network Flow by expanding the state space.\\
\textbf{Core Strategy:}
\begin{enumerate}
    \item \textbf{Redefine} the state: move from simple states (e.g., `dp[i]`) to composite, multi-dimensional states (e.g., adding an exponential `mask` for sets or a polynomial `remainder` for constraints).
    \item \textbf{Introduce} constraints that depend on past choices or specific history (e.g., "cannot visit a node visited k steps ago," requiring a sliding window or history state).
    \item \textbf{Add} multiple orthogonal restrictions (e.g., count, parity, or modularity) that must be tracked simultaneously.
\end{enumerate}

\textbf{Anti-pattern:} Simply adding variables that don't interact. The new dimensions must fundamentally change the recurrence logic or transition dependencies.\\
\textbf{Examples:}
\begin{lstlisting}
[
    {
        "original": "Climb stairs taking 1 or 2 steps. How many ways?",
        "upgrade": "Cover a 3×N grid with 1×2 dominoes. How many ways modulo 10^9+7? (Requires Broken Profile DP / Bitmask DP to track cross-section state)"
    },
    {
        "original": "Knapsack Problem: Max value with weight limit W.",
        "upgrade": "Knapsack on a Tree: Each node has value/weight. Max value by selecting nodes such that no two selected nodes are adjacent, and total weight ≤ W. (Requires Tree DP + Knapsack dimensions)"
    },
    {
        "original": "Longest Increasing Subsequence in an array.",
        "upgrade": "Count the number of permutations of length N that have a Longest Increasing Subsequence of length exactly K. (Requires DP with Young Tableaux or RSK correspondence)"
    },
    {
        "original": "Find min cost to traverse a grid.",
        "upgrade": "Find min cost to traverse a grid with K 'batteries' to jump obstacles; recharge depends on grid value modulo M, and cells must be unlocked in order. (State: position + battery_count + mod_state + lock_mask)"
    },
    {
        "original": "Edit Distance between two strings.",
        "upgrade": "Given strings A and B, find the number of strings S of length L such that EditDistance(A, S) ≤ K and EditDistance(B, S) ≤ K. (Requires DP on DP / Automaton DP)"
    }
]
\end{lstlisting}

% 5
\textbf{\#\#\# Category E: }\\
\textbf{Theorem Disguise}\\
\textbf{Best for:} Problems that can map to classic high-level algorithms but appear in unrelated or abstract domains.\\
\textbf{Avoid when:} The original problem explicitly mentions the algorithm or data structure.\\
\textbf{Description:} Hide a sophisticated algorithmic core behind a narrative or alternative mathematical structure that misleads intuition.\\
\textbf{Core Strategy:}
\begin{enumerate}
    \item \textbf{Map} the problem to a well-known non-trivial algorithm (e.g., Network Flow, Linear Basis, Generating Functions, or Advanced DS).
    \item \textbf{Remove} all technical terminology and explicit constraints that hint at the solution.
    \item \textbf{Create} a "Red Herring" narrative that suggests an intuitive but suboptimal approach (e.g., Greedy, simple DP, or naive Simulation).
    \item \textbf{Ensure} the bridge between the surface problem and the hidden theorem requires a deep structural insight.
\end{enumerate}
\textbf{Examples:}
\begin{lstlisting}
[
    {
        "original": "Find the maximum number of meetings one can attend.",
        "upgrade": "Given a set of intervals, color them with minimum colors so no overlapping intervals share a color. Some intervals are 'VIP' and must use specific colors. (Disguised as Greedy, actually Interval Graph Coloring / Dilworth's Theorem with constraints)"
    },
    {
        "original": "Can we partition an array into two equal sum subsets?",
        "upgrade": "Given a graph, determine if it is bipartite. If not, remove minimum edges to make it bipartite while preserving connectivity. (Disguised Graph Theory, actually Odd Cycle Transversal / Max Cut variant)"
    },
    {
        "original": "Assign tasks to workers to minimize total time.",
        "upgrade": "N workers and M tasks. Each pair (worker, task) has a cost. Some workers conflict and cannot work simultaneously. Minimize total cost. (Disguised Minimum Cost Maximum Flow with conflict resolution)"
    },
    {
        "original": "Find the maximum number of compatible intervals.",
        "upgrade": "Given a circle with N chords represented by their endpoints. Find the maximum number of non-intersecting chords. (Disguised DP on Circle Graph or Catalan-related structure)"
    },
    {
        "original": "Find a number in an array that appears more than N/2 times.",
        "upgrade": "Given a stream of N elements, find all elements appearing more than N/K times using O(K) space and one pass. Handle K up to 10^3. (Disguised Boyer-Moore Voting generalized / Misra-Gries Algorithm)"
    }
]
\end{lstlisting}

% 6
\textbf{\#\#\# Category F: }\\
\textbf{Edge Case \& Rigor Engineering}\\
\textbf{Best for:} Problems with simple logic but tricky implementation details.\\
\textbf{Avoid when:} The original problem is already highly technical or implementation-heavy.\\
\textbf{Description:} Focus on logical pitfalls, geometric precision, tricky edge cases, or requiring formal proof for correctness.\\
\textbf{Core Strategy:}
\begin{enumerate}
    \item \textbf{Target} weaknesses of standard data types (e.g., float precision, integer overflow, empty sets).
    \item \textbf{Introduce} degenerate cases (e.g., disconnected graphs, collinear points, zero denominators).
    \item \textbf{Require} strict mathematical proofs (like "Exchange Arguments") to justify a strategy where intuition fails.
    \item \textbf{Add} precision constraints (e.g., "answer accurate to $10^{-9}$") or handle extreme ranges.
\end{enumerate}
\textbf{Examples:}
\begin{lstlisting}
[
    {
        "original": "Given coordinates of 3 points, calculate the area of the triangle.",
        "upgrade": "Given N (N≤10^5) lines in a plane. Find the area of the finite region formed by their intersection. Handle parallel lines, lines at infinity, and precision errors. (Requires Half-plane Intersection with strict boundary handling and epsilon comparison)"
    },
    {
        "original": "Find if there is a path from A to B in a graph.",
        "upgrade": "Find if A reaches B. The graph may contain self-loops, multiple edges, and consists of up to 10^5 disconnected components. Answer 10^5 online queries. Handle the case where A=B and the case where A or B doesn't exist. (Tests Edge Cases: Connectivity, Disjoint Set Union with path compression, query validation)"
    },
    {
        "original": "Divide A by B.",
        "upgrade": "Divide A by B where A and B are strings representing integers up to 10^1000. Handle B=0, negative signs, leading zeros, and output the quotient and remainder. Prove your division algorithm's correctness. (Requires BigInteger Implementation + comprehensive Edge Case handling)"
    },
    {
        "original": "Greedy: Always pick the largest coin available to make change.",
        "upgrade": "Given a non-standard coin system (e.g., 1, 3, 4), find the smallest value V where the greedy approach fails to find the minimum number of coins. Prove that for all values below V, greedy is optimal. (Requires finding the Edge Case of the algorithm itself + Exchange Argument)"
    },
    {
        "original": "Sort an array of tasks by duration.",
        "upgrade": "Given N tasks with processing time and decay rates. Each task loses value at a different rate while waiting. Find a processing order to minimize total decay. Prove your sorting strategy using an Exchange Argument. (Requires formal proof to derive the correct comparator; simple sorting is wrong)"
    }
]
\end{lstlisting}

\textbf{\#\# Construction Protocol (Internal Thinking Process)}
\begin{enumerate}
    \item \textbf{Analyze Original}: Identify the naive solution and its complexity.
    \item \textbf{Select Category}: Choose 1-2 categories from the library above that best fit the problem's potential.
    \item \textbf{Apply Core Strategy}: Use the "Core Strategy" defined in your selected category to redesign the problem constraints and objectives.
    \item \textbf{Review}: Check against the "Design Standards". Does it require multi-step reasoning? Is the theorem disguised?
    \item \textbf{Final Output}: Write the problem statement clearly using standard CP formatting.
\end{enumerate}

\textbf{\#\# Input}

\textbf{Original Problem:}\\
\{\text{original\_problem}\}\\

\textbf{\#\# Output Format}\\
Return ONLY the new upgraded problem, nothing else.\\

[Your upgraded competitive programming problem here]

\end{promptbox}

\end{document}